\documentclass{article}

\PassOptionsToPackage{numbers, compress}{natbib}
\usepackage{arxiv}
\usepackage[utf8]{inputenc}
\usepackage[T1]{fontenc}
\usepackage{url}
\usepackage{booktabs}
\usepackage{amsfonts}
\usepackage{nicefrac}
\usepackage{microtype}
\usepackage{xcolor}
\usepackage{subcaption}
\usepackage{epsfig}
\usepackage{float}
\usepackage{xspace}
\usepackage{amsmath}
\usepackage{amssymb}
\usepackage{bm}
\usepackage{array}
\usepackage{algorithm}
\usepackage{algpseudocode}
\usepackage{upgreek}
\usepackage{bbm}
\usepackage{mathtools}
\usepackage{color}
\usepackage{multirow}
\usepackage{enumitem}
\usepackage{wrapfig}

\makeatletter
\DeclareRobustCommand\onedot{\futurelet\@let@token\@onedot}
\def\@onedot{\ifx\@let@token.\else.\null\fi\xspace}

\makeatother

\DeclareMathOperator*{\argmax}{argmax}

\DeclareMathOperator*{\similarity}{sim}
\definecolor{canaryyellow}{rgb}{1.0, 0.94, 0.0}
\definecolor{corn}{rgb}{0.98, 0.93, 0.36}
\definecolor{hansayellow}{rgb}{0.91, 0.84, 0.42}
\definecolor{darkyellow}{rgb}{1.0, 0.83, 0.0}

\newcommand{\name}{\textsc{reprogrammer}}
\newcommand{\resname}{\textsc{residual reprogrammer}}
\newcommand\blfootnote[1]{%
  \begingroup
  \renewcommand\thefootnote{}\footnote{#1}%
  \addtocounter{footnote}{-1}%
  \endgroup
}

\title{Model Reprogramming Outperforms Fine-tuning on Out-of-distribution Data in Text-Image Encoders}

\usepackage{authblk}

\author[1, 2]{\textbf{Andrew Geng}}
\author[1]{\textbf{Pin-Yu Chen}}

\affil[1]{\footnotesize \textit{IBM Research, Yorktown Heights, NY}}
\affil[2]{\footnotesize \textit{University of Wisconsin-Madison, Madison, WI}}
\affil[ ]{\{andrew.geng, pin-yu.chen\}@ibm.com}

\begin{document}

\maketitle

\begin{abstract}
When evaluating the performance of a pre-trained model transferred to a downstream task, it is imperative to assess not only the in-distribution (ID) accuracy of the downstream model but also its capacity to generalize and identify out-of-distribution (OOD) samples. In this paper, we unveil the hidden costs associated with intrusive fine-tuning techniques. Specifically, we demonstrate that commonly used fine-tuning methods not only distort the representations necessary for generalizing to covariate-shifted OOD samples (OOD generalization) but also distort the representations necessary for detecting semantically-shifted OOD samples (OOD detection). To address these challenges, we introduce a new model reprogramming approach for fine-tuning, which we name \name{}. \name{} aims to improve the holistic performance of the downstream model across ID, OOD generalization, and OOD detection tasks. Our empirical evidence reveals that \name{} is less intrusive and yields superior downstream models. Furthermore, we demonstrate that by appending an additional representation residual connection to \name{}, we can further preserve pre-training representations, resulting in an even more safe and robust downstream model capable of excelling in many ID classification, OOD generalization, and OOD detection settings.\blfootnote{Code is made publicly available at https://github.com/IBM/reprogrammer.} 
\end{abstract}

\section{Introduction}
\label{sec:intro}
As pre-trained models become increasingly adopted for addressing complex downstream tasks, it has become progressively more important to ensure not only the in-distribution accuracy of the downstream model but also its robustness and safety when confronted with distribution shifts. In real-world applications, models often encounter samples that deviate to varying degrees from the expected in-distribution dataset. For samples exhibiting covariate shifts (non-semantic) from the in-distribution, we assess robustness by measuring the OOD generalization, where a robust model should consistently maintain high accuracy across all covariate-shifted OOD samples. Alternatively, for samples exhibiting semantic shifts from the in-distribution, we evaluate safety through OOD detection, where a safe and robust model should be capable of distinguishing semantically shifted OOD samples from the ID samples. Recently, both of these problems have been rigorously studied with a plethora of new and exciting literature aimed at addressing these issues~\cite{chen2021robustifying, hsu2020generalized, huang2021mos, kumar2022, lakshminarayanan2017simple, lee2018simple, liang2018enhancing, lin2021mood, liu2020energy, miller2021, mohseni2020self, nalisnick2018deep, radford2021, taori2020, wortsman2021}.

However, several fundamental challenges still impede researchers from improving ID, OOD generalization, and OOD detection performances. These challenges range from difficulties in encapsulating covariant (domain) shifts, to overconfidence when predicting semantically shifted samples \cite{miller2021, taori2020, nguyen2015deep}. One framework, for training deep learning models, that has demonstrated strong performance in both ID classification and OOD generalization settings is large text-image supervised pre-trained models~\cite{jia2021, pham2021, radford2021}. However, it has recently become apparent that common fine-tuning methods can distort the robust representations acquired during multi-modal pre-training, which can result in a decline in the fine-tuned model's OOD generalization performance \cite{andreassen2021, kumar2022, wortsman2021}. Moreover, it also remains unclear whether these distortions, induced by fine-tuning, will adversely affect OOD detection tasks in the same way observed in OOD generalization.

In this paper, we present evidence demonstrating that common fine-tuning techniques, such as linear-probing (optimizing only the classification head), full fine-tuning (optimizing all model parameters), LP-FT (optimizing classification head first before full fine-tuning) \cite{kumar2022}, regularized fine-tuning (optimizing all model parameters while applying regularization to the zero-shot weights) \cite{Li2018LPSP}, model soups (average the weights of zero-shot and fine-tuned models) \cite{wortsman2021}, and prompt learning (optimizing adjustable tokens in the caption) \cite{zhou2022coop}, can not only degrade OOD generalization performance but also compromise OOD detection capabilities. Furthermore, we illustrate that each of these common fine-tuning techniques possesses distinct strengths and hidden costs associated with their ID, OOD generalization, and OOD detection capabilities. This raises the question \textit{can we develop an alternative fine-tuning technique that is less intrusive, more robust, and safer on both covariate and semantically shifted OOD samples?}

\begin{figure*}[t]
    \captionsetup[subfigure]{justification=centering}
    \centering
    \begin{subfigure}{0.245\textwidth}
        \centering
        \includegraphics[width=\textwidth]{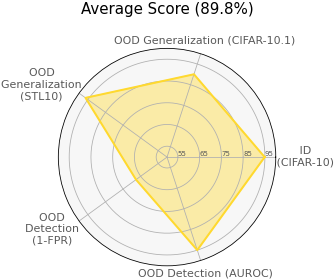}
        \caption{Linear-probing}
    \end{subfigure}
    \begin{subfigure}{0.245\textwidth}
        \centering
        \includegraphics[width=\textwidth]{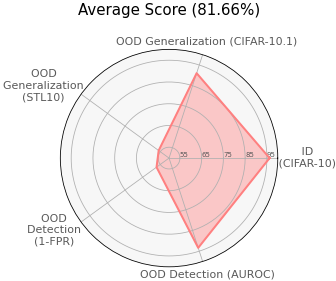}
        \caption{Full Fine-tuning}
    \end{subfigure}
    \begin{subfigure}{0.245\textwidth}
        \centering
        \includegraphics[width=\textwidth]{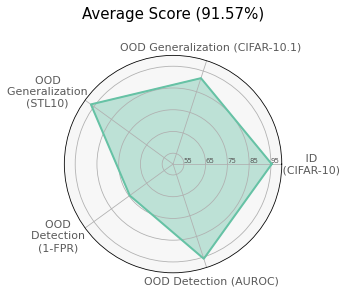}
        \caption{Reprogrammer}
    \end{subfigure}
        \begin{subfigure}{0.245\textwidth}
        \centering
        \includegraphics[width=\textwidth]{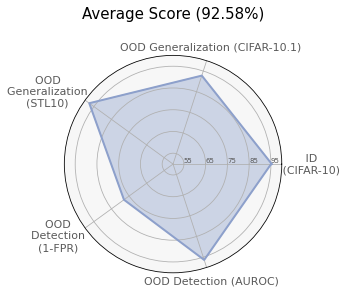}
        \caption{Residual Reprogrammer}
    \end{subfigure}
    \caption{Radar charts illustrating the trade-offs between ID, OOD generalization, and OOD detection performances across linear-probing, full fine-tuning, \name{}, and \resname{}. All results are based on the CIFAR benchmarks. To quantify the cost-performance trade-offs, we report the average scores normalized across all metrics.}
    \label{fig:teaser}
    \vspace{-0.2cm}
\end{figure*}

We tackle this question by exploring and altering an alternative approach to transfer learning called \textit{model reprogramming} \cite{chen2022}. By leveraging and altering key components of \textit{model reprogramming}, we present a new method for reprogramming a text-image pre-trained model to a downstream ID task. We also show that due to the less intrusive nature of \textit{model reprogramming} (no adjustments to the pre-trained model parameter) our approach is less distortive, leading to improved OOD generalization and OOD detection performances. In addition, our findings also reveal that by incorporating a representation residual connection into \name{}, we can further promote the retention of pre-training representations. Our operating hypotheses are
\vspace{0.1cm}
\begin{itemize}[leftmargin=0.9cm]
  \itemsep0em
  \item[\textbf{[H1]}] Traditional fine-tuning techniques can degrade both OOD generalization and OOD detection performances in CLIP-like models, leading to worse OOD performance when compared to the untuned zero-shot model.
  \item[\textbf{[H2]}] By solely employing \textit{model reprogramming} techniques, which are less intrusive and do not impose any changes to the pre-trained model parameters, our \name{} will lead to more pre-training representations being maintained throughout the fine-tuning process.
  \item[\textbf{[H3]}] The addition of a representation residual connection to the zero-shot model can further maintain pre-training representations, leading to enhanced OOD generalization and detection performances.
  \item[\textbf{[H4]}] Reprogramming the image encoder on ID samples enables \name{} to more effectively align covariate-shifted OOD samples with the in-distribution space during inference, consequently resulting in enhanced ID classification and OOD generalization.
\end{itemize}

More formally, we introduce \name{} and \resname{}, a pair of \textit{model reprogramming} techniques that leverage two distinct modalities of reprogramming functions to simultaneously reprogram the image encoder and the text encoder. Subsequently, we conduct a comprehensive set of evaluations demonstrating the superiority of our \name{} and \resname{} methods. To the best of our knowledge, we are the first to venture into applying \textit{model reprogramming} techniques to multi-modal joint text-image encoder models. An illustration depicting the trade-offs between cost and performance associated with intrusive fine-tuning techniques can be found in Figure~\ref{fig:teaser}. 

Our \textbf{key results and contributions} are summarized as follows:
\begin{itemize}[leftmargin=*]
  \itemsep0em
  \item We demonstrate that common fine-tuning techniques can degrade OOD performances, resulting in trade-offs between ID, OOD generalization, and OOD detection. A visual comparison of these trade-offs in terms of cost and performance is provided in Figure \ref{fig:teaser}.
  \item We introduce \name{} and \resname{}, a pair of simple yet effective fine-tuning techniques designed to fully maintain and harness pre-training representations in CLIP-like models.
  \item Our results show that \resname{} consistently outperforms all other methods holistically when evaluating ID, OOD generalization, and OOD detection tasks. Improving the aggregated performance by $+2.78\%$ on CIFAR benchmarks and $+0.69\%$ on \texttt{ImageNet-1k} benchmarks when compared to the next best method.
  \item Additionally, we conduct supporting ablations to improve our understanding of \name{} under (1) varying degrees of reprogramming strength and (2) visualizing the reprogrammed feature space under covariate shifts.
\end{itemize}

\section{Background and Related Work}
\label{sec:background}

\paragraph{Pre-trained and CLIP-like Models:}
Pre-trained models, trained on vast and diverse datasets, have become a popular technique for constructing robust machine learning models capable of efficient transfer to downstream tasks \cite{brown2020, chen2020recall, desai2021, dosovitskiy2020, chao2021, kolesnikov2020, radford2019, mert2020, xiaohua2021, zhang2020}. In this paper, we primarily focus on the Contrastive Language-Image Pre-training (CLIP) model \cite{radford2021}. CLIP is a multi-modal model pre-trained on a large dataset of 400 million image-caption pairs collected from the web. More specifically, given a set of image-caption pairs $D= \{(X_1, T_1)...,(X_n, T_n)\}$, CLIP-like models train an image-encoder $f$ and a text-encoder $g$ such that the cosine similarity between the features $f(x_k)$ and $h(t_k)$ are maximized with respect to each pair $k$.

\paragraph{Out-of-distribution Generalization:}
To assess the OOD generalization performance of our downstream models, we fine-tune and compare the accuracy of our tuned models using two distinct yet interconnected datasets $D_{in}$ and $D_{out}$. The dataset $D_{in}$ corresponds to the in-distribution dataset to which our pre-trained model is tuned on. The OOD dataset $D_{out}$ represents a covariate (domain) shifted out-of-distribution dataset, comprising samples that share the same semantic classifications as those in the in-distribution dataset $D_{in}$ but manifested under different domains. These domains can include sketches, origami, and other variations of the in-distribution classes \cite{hendrycks2021many, hendrycks2021nae, recht2019, wang2019learning}.

In an OOD generalization context, the goal of an effective fine-tuning technique is to attain high accuracy across both $D_{in}$ and $D_{out}$. Being able to achieve high accuracy across both datasets is paramount, as an intelligent and robust model should be agnostic to the covariate shifts of a given sample.

\paragraph{Out-of-distribution Detection:}
Out-of-distribution detection can be formulated as a binary classification problem where, given some classifier $\tilde{f}$ tasked on the in-distribution dataset $D_{in}$, our objective is to design a function estimator
\begin{equation*}
    h(\hat{x}) = 
    \begin{cases}
    \text{in}, &\text{if}\ S(\hat{x}) \geq \gamma \\
    \text{out}, &\text{if}\ S(\hat{x}) < \gamma,
    \end{cases}
\end{equation*}
such that $h(\hat{x})$ can determine whether a sample $\hat{x}$ is in-distribution $D_{in}$ or out-of-distribution $Q_{out}$. 

Critically, in the OOD detection setting, our goal is to detect semantically shifted samples. For instance, if the in-distribution encapsulates samples of $\{``\textit{cats}", ``\textit{dogs}"\}$ then the goal of our detector $h$, given a $``\textit{car}"$ sample $\hat{x}$, is to detect that the $\hat{x}$ sample does not belong to the in-distribution set $\hat{x} \notin D_{in}$, or equivalently that the sample is out-of-distribution $\hat{x} \in Q_{out}$. To evaluate OOD detection, we employ the commonly used maximum softmax probability (\texttt{msp}) detector $h_{msp}$ \cite{hendrycks2016baseline}, which measures the confidence of our classifier $\tilde{f}$ towards a given input $\hat{x}$. The goal, of a strong fine-tuning method, is to produce a downstream model $\tilde{f}$ that is more uncertainty aware. Specifically, we want the downstream model $\tilde{f}$ to not confidently classify on semantically shifted OOD samples, whilst maintaining confidence when predicting ID samples. This goal is again immediately apparent, as we want a safe and robust model to not (overconfidently) find a semantically dissociated OOD sample to be indistinguishable from ID samples \cite{hendrycks2016baseline, hsu2020generalized, liang2018enhancing, liu2020energy}.

\paragraph{Model Reprogramming:}
\textit{Model reprogramming} is a resource-efficient, cross-domain, framework used to re-purpose models for different task-specific scenarios~\cite{chen2022}. The framework draws significant inspiration from adversarial reprogramming, which was first introduced by Elsayed et al~\cite{elsayed2019}. The aim of \textit{model reprogramming} is to re-use and re-align the data representation, from an existing model, for a separate task without fundamental changes to the model's parameters \cite{yang2021voice2series}. 
In particular, \textit{model reprogramming} utilizes a trainable input transformation (reprogramming function) that maps the input to a new form for the model to ingest. Following a forward pass, \textit{model reprogramming} employs a label mapping function to generate final classification predictions. It is important to note that reprogramming functions are not specific to any singular input; instead, the reprogramming function is consistently applied to all inputs.
Traditionally, \textit{model reprogramming} methods operate by training an image/audio reprogramming function to optimally transform continuous input data, such that the output of the model can be used to perform some other desired task \cite{elsayed2019,yang2021voice2series}.  \textit{Model reprogramming} methods have also been proven to be successful in both white-box and black-box settings \cite{tsai2020}. Additionally, Neekhara et al~\cite{neekhara2019} presented a reprogramming method for sequence classification models, by utilizing a context-based vocabulary remapping function \cite{neekhara2019, neekhara2022}. To the best of our knowledge, this paper is the first \textit{model reprogramming} method tackling joint text-image encoders in a multi-modal setting.

\begin{figure*}[t]
    \captionsetup[subfigure]{justification=centering}
    \centering
    \begin{subfigure}{0.98\textwidth}
        \centering
        \includegraphics[width=\textwidth]{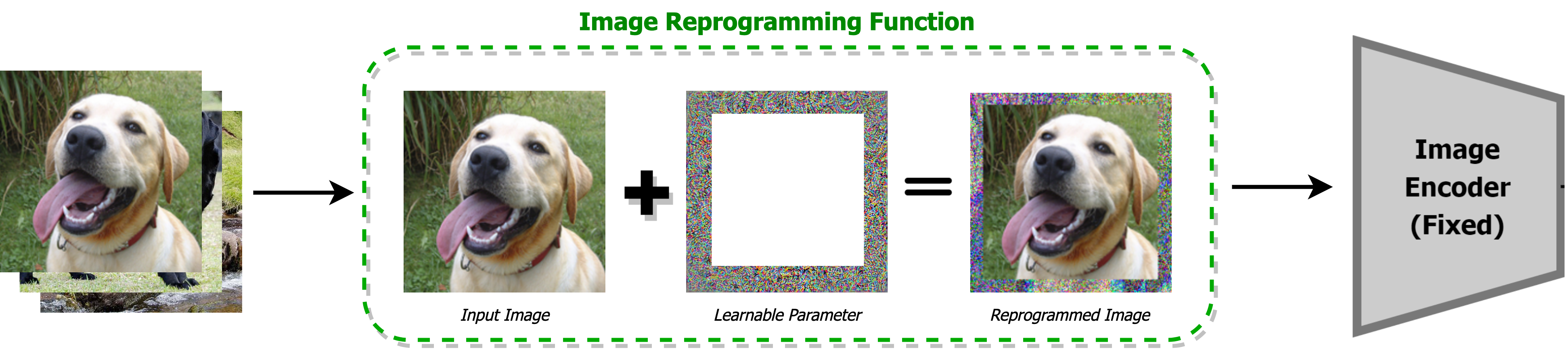}
    \end{subfigure}
    \begin{subfigure}{0.98\textwidth}
        \centering
        \includegraphics[width=\textwidth]{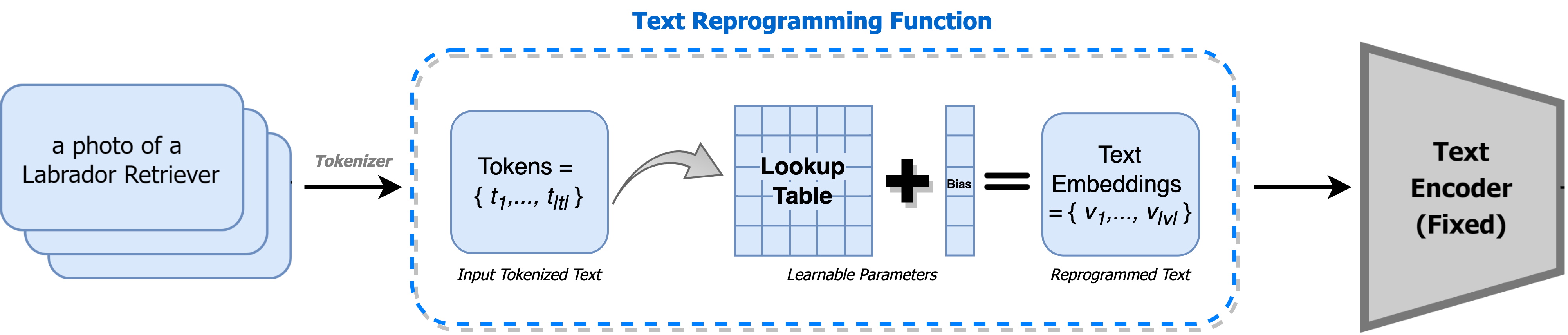}
    \end{subfigure}
    \caption{Visual diagrams illustrating the image reprogramming and text reprogramming functions. In the image reprogramming function, an input image undergoes resizing and padding, followed by the addition of a learnable edge perturbation. Similarly, in the text reprogramming function, an input caption is tokenized before a lookup table and bias embedding are applied. Subsequently, both the reprogrammed image and caption embeddings are passed through the fixed text-image encoder during a model forward pass.}
    \label{app:reprogrammer_methodology}
    \vspace{-0.2cm}
\end{figure*}

\section{Methodology}
\label{sec:method}
In this section, we begin by introducing the image and text reprogramming modules as presented in Figure~\ref{app:reprogrammer_methodology}, followed by the full \name{} and \resname{} fine-tuning techniques. Additionally, we offer further details on our methodology in Appendix~\ref{app:detail_method}.

\subsection{Image Reprogramming}
Consider just the CLIP image encoder $f:I \rightarrow \mathbb{R}^{b\times k}$ where $b$ is the input image batch size and $k=512$ is the CLIP feature size. To apply reprogramming, we leverage the commonly used adversarial program first described by Elsayed et al~\cite{elsayed2019}, which we define as the reprogramming function $\psi$. The reprogramming function $\psi$ is applied to the input image pre-forward pass through the CLIP image encoder $f$. Critically, the reprogramming function $\psi$ is not specific to any singular input image, rather $\psi$ will be consistently applied to all images. We define our reprogramming function $\psi$ as
\begin{align}
\psi(X) = \mathcal{U}(X) + \tanh (W \odot M)
\end{align} 
where $\mathcal{U}$ denotes an image up-sampling then zero-padding function, $W\in \mathbb{R}^{d \times d \times 3}$ is the image reprogrammer parameters that is to be learned, $d$ is the size of CLIP's input width and height, $\odot$ denotes the Hadamard product, and $M$ is a binary masking matrix. We define the binary masking matrix $M$ as $0$ for positions where we wish to implant the original image, and $1$ for positions that we choose to reprogram.


\subsection{Text Reprogramming}
Now we consider the CLIP text encoder $g: S\rightarrow \mathbb{R}^{b\times k}$ where $b$ is the input text batch size and $k=512$ is the CLIP feature size. Additionally, we define our text input $s$ as a sequence of tokens $s =\{s_1,..., s_{|s|}\}$ where $s_i$ is the vocabulary index of the $i^{th}$ token in the vocabulary list $V_S$. To apply reprogramming to a text input, we leverage and alter a version of the adversarial program as first described by Neekhara et al~\cite{neekhara2019}. 

Formally, we define our text reprogramming function as $\Phi_{\theta,b}$ where $\Phi_{\theta,b}$ is a simple look-up embedding and bias on the tokens $\{s_i\}$ that can be parameterized by the learnable embedding tensor $\theta$ and the bias parameter $b$. Specifically, we define our $\theta \in \mathbb{R}^{|V_S|\times d}$ and $b \in \mathbb{R}^{d}$ where our default vocabulary size is $|V_S| = 49408$, which is the expected vocabulary size for the CLIP text encoder. Similarly, as with all reprogramming functions, the text reprogramming function is not specific to any singular text input, rather $\Phi_{\theta,b}$ will be consistently applied to all text inputs. 

An example of the text reprogramming function goes as follows. First, we set $s_i = ``\text{a photo of a }\{c_i\} "$ where $c_i$ is the given sample class label. Our text reprogramming then tokenizes the string $s = ``\text{a photo of a Labrador Retriever}"$ into tokens $t_s$. Subsequently, the tokens $t_s$ are passed into the $\Phi_\theta$ function to embed the tokens into a matrix $v'_s\in \mathbb{R}^{|t_s|\times e}$ where each token in $t_s$ becomes an embedding vector of size $e$. Then we apply a bias parameter $b$ to $v'_s$ in the form of $v_s = v'_s + b$, before finally passing the vector $v_s$ through the CLIP text encoder $g$ to get the reprogrammed text features.

\begin{figure*}[t]
    \captionsetup[subfigure]{justification=centering}
    \centering
    \begin{subfigure}{0.98\textwidth}
        \centering
        \includegraphics[width=\textwidth]{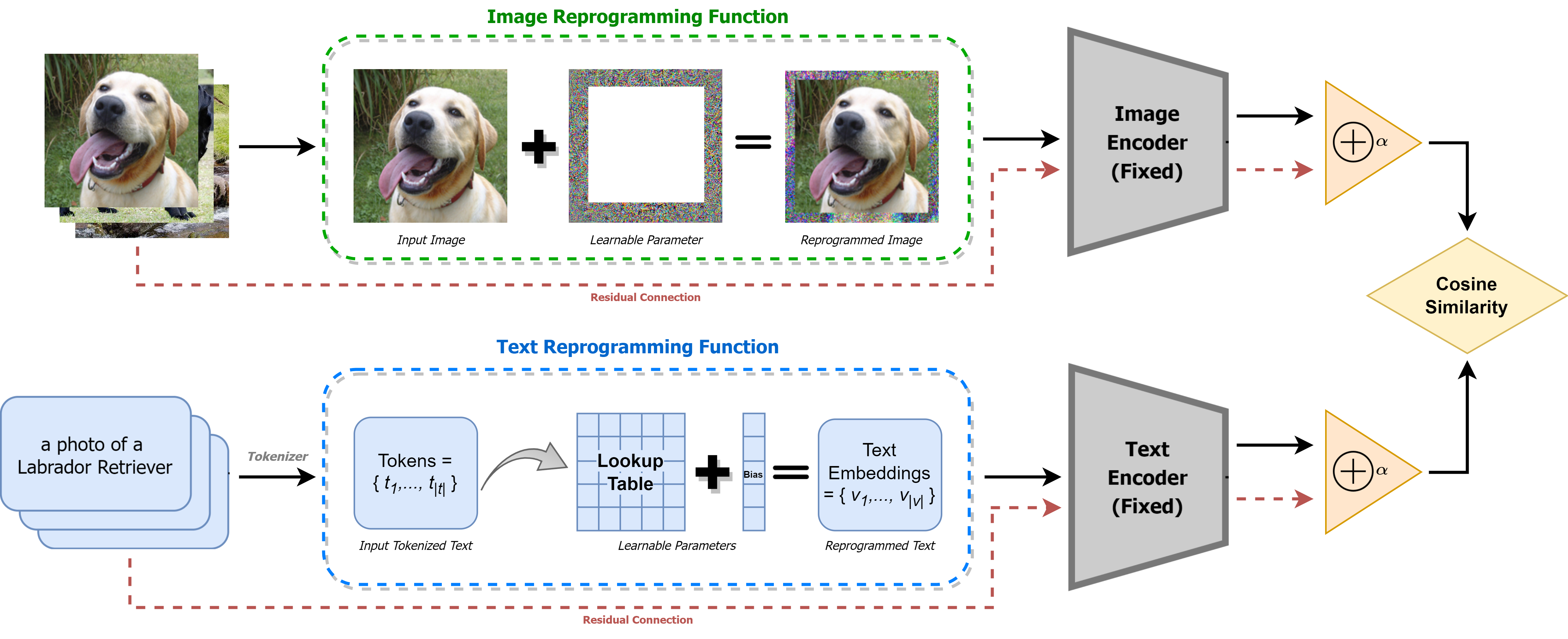}
    \end{subfigure}
    \caption{Visual diagram illustrating the \name{} and \resname{} training schema based on the CLIP joint image and text encoder setting. During \name{} training, an image and caption pair each independently undergoes their respective reprogramming functions before being passed into the CLIP image and text encoders. A loss is then computed based on the cosine similarity of the two reprogrammed features. Then we subsequently backpropagate and optimize each parameter associated with the image and text reprogramming function. During inference time, \resname{} leverages a residual connection that combines the reprogrammed representation and zero-shot representations.}
    \label{fig:image_reprogrammer_methodology}
    \vspace{-0.2cm}
\end{figure*}

\subsection{Reprogrammer}
Finally, to train our given image and text reprogramming functions $\psi$ and $\Phi_{\theta,b}$, we define our training objective as
\vspace{0.1cm}
\begin{align}
W^*, \theta^*, b^* = \argmax_{W, \theta, b}\left( \similarity(f( \psi_W(x) ), \; g(\Phi_{\theta,b}(s)) ) \right)
\end{align}
where $(x,s)$ is an image and caption pair obtained from our training set $D_{in}$, $f$ and $g$ are the CLIP image and text encoders respectively, $\similarity$ is the cosine-similarity function, and $W, \theta, b$ are the learnable parameters encapsulating our reprogramming functions $\psi_W, \Phi_{\theta,b}$. In practice, rather than directly optimizing for cosine similarity, we follow the optimization schema of a symmetric cross-entropy loss as implemented in CLIP pre-training \cite{radford2021}. It is important to note that throughout the \name{} training process, we impose no adjustments to any pre-trained model parameters. Thereby fundamentally limiting any distortion to the pre-training representations.

After tuning our \name{} parameters $W, \theta, b$ we perform classification, on an input image $\hat{x}$ with $m$ class labels $C=\{c_1, ..., c_m\}$, similar to that of zero-shot CLIP. Specifically, we make a prediction $y$ through
\vspace{0.1cm}
\begin{align}
y=\argmax_i (\similarity(f( \psi_{W^*}(\hat{x}) ), g(\Phi_{\theta^*, b^*}(s_i))
\end{align}
where $s_i$ is the class-wise captions such that $s_i = ``\text{a photo of a }\{c_i\} "$ and $\psi_{W^*}$ and $\Phi_{\theta^*, b^*}$ are our learned reprogramming functions parameterized by $W^*$, $\theta^*$, and $b^*$.

\subsection{Residual Reprogrammer}
To further retain pre-training representations, we propose \resname{} which seeks to fuse prior pre-training representations alongside our reprogrammed representations. These representation residual connections have been utilized in different formulations and settings \cite{peng2021, zhang2022tip} however they have never been applied in the context of \textit{model reprogramming}. 
Additionally, it is important to note that the representation residual connections employed in \resname{} are not the residual connections defined in the ResNet architecture \cite{he2016deep}. 
Furthermore, we provide some intuition in Appendix~\ref{app:analysis} showcasing how \resname{} can be interpreted as an inference time regularizer for our image and text reprogramming functions.

Consider a tuned \name{} model with parameters $W, \theta, b$ and an input image $\hat{x}$ with $m$ class labels $\{c_1, ..., c_m\}$. We define our residual reprogramming functions as
\begin{align}
F(\hat{x}) &= (1-\alpha)f( \psi_{W^*}(\hat{x}) ) + \alpha f(\hat{x})\\
G(s_i) &= (1-\alpha)g(\Phi_{\theta^*, b^*}(s_i)) + \alpha g(s_i)
\end{align}
where $\psi_{W^*}$ and $\Phi_{\theta^*, b^*}$ are our learned reprogramming functions parameterized by $W^*$, $\theta^*$, and $b^*$. Subsequently, during inference time, we perform classification with \resname{} through
\begin{align}
y=\argmax_i (\similarity(F(\hat{x}), G(s_i))
\end{align}
where $s_i$ is the class-wise captions such that $s_i = ``\text{a photo of a }\{c_i\} "$. 

\section{Experiments}
\label{sec:experiments}
In this section, we first outline our experimental setup for OOD generalization and OOD detection in Section~\ref{sec:exp_setup}, before evaluating our \name{} and \resname{} methods against other common fine-tuning techniques in Section~\ref{sec:results}. Furthermore, we conduct additional ablations in Section~\ref{sec:ablations}. More experimental details and supplementary studies can be found in Appendix~\ref{app:exp_detail} and Appendix~\ref{app:additional_experiments}.

\subsection{Experimental Setup}
\label{sec:exp_setup}

\paragraph{In-distribution dataset:} 
We tune our model with \texttt{CIFAR-10} \cite{krizhevsky2009learning} and \texttt{ImageNet-1k} \cite{deng2009imagenet} as the in-distribution (ID) datasets. These datasets are widely employed as ID datasets for OOD generalization and OOD detection experiments. The \texttt{CIFAR-10} dataset contains labeled (32$\times$32) resolution images covering a range of real-world objects such as horses, cats, and airplanes. The ImagetNet-1k dataset contains over 1.2 million training images spanning 1000 different real-world objects such as species of dogs and automotive vehicles.

\paragraph{Out-of-distribution Generalization:}
For models fine-tuned on \texttt{CIFAR-10}, we evaluate the OOD generalization performance on two standard covariate-shifted OOD datasets. Specifically, we evaluate generalization accuracy with the \texttt{CIFAR-10.1} \cite{torralba2008} and \texttt{STL10} \cite{coates2011} datasets. For models fine-tuned with \texttt{ImageNet-1k}, we evaluate the OOD generalization performance across four widely used benchmarks. In particular, we evaluate generalization accuracy with \texttt{ImageNetV2} \cite{recht2019}, \texttt{ImageNet-R} \cite{hendrycks2021many}, \texttt{ImageNet-A} \cite{hendrycks2021nae}, and \texttt{ImageNet-Sketch} \cite{wang2019learning}. All of these datasets contain images derived from the same semantic labels as those in the ID dataset. For example, these datasets may encompass a sketched version of a Labrador Retriever, a cartoon depiction of a strawberry, or a photograph of a toy duck (for more details refer to Appendix~\ref{app:datasets}).

\paragraph{Out-of-distribution Detection:}
For models fine-tuned on \texttt{CIFAR-10}, we evaluate using the \texttt{msp} detector against four commonly used CIFAR OOD detection benchmarks. More specifically, we evaluate on the \texttt{iSUN}~\cite{pingmei2015}, \texttt{LSUN Resized}~\cite{yu15lsun}, \texttt{Places365}~\cite{zhou2017places}, and \texttt{Textures}~\cite{cimpoi14describing} datasets. These OOD datasets span a wide range of objects including fine-grained images, scene images, and textural images. Importantly, these datasets are carefully chosen so that there is no semantic overlapping with respect to the \texttt{CIFAR-10} dataset. For models tuned with \texttt{ImageNet-1k}, we use the large-scale ImageNet OOD detection benchmark proposed by Huang et al~\cite{huang2021mos}. Specifically, we evaluate on four OOD datasets which are subsets from the \texttt{iNaturalist} \cite{van2018inaturalist}, \texttt{SUN} \cite{xiao2010sun}, \texttt{Places} \cite{zhou2017places}, and \texttt{Textures} \cite{cimpoi14describing} datasets. These datasets are again carefully curated so that there is no semantic overlap with respect to the \texttt{ImageNet-1k} dataset (for more details refer to Appendix~\ref{app:datasets}).

\begin{table*}[t]
\centering
\setlength\tabcolsep{7pt} 
\renewcommand{\arraystretch}{1.1}
\scriptsize{
\begin{tabular}{c|c|c|c|c|c|c}
\toprule
\multicolumn{1}{c|}{\multirow{2}{*}{\textbf{$\bm{D_{in}}$}}} & \multirow{2}{*}{\textbf{\begin{tabular}[c]{@{}c@{}} Method \end{tabular}}} & \multicolumn{1}{c|}{\textbf{ImageNet-1k}} & \multicolumn{1}{c|}{\textbf{ImageNetV2}} & \multicolumn{1}{c|}{\textbf{ImageNet-A}} & \multicolumn{1}{c|}{\textbf{ImageNet-R}} & \multicolumn{1}{c}{\textbf{ImageNet-S}} \\ 
\cline{3-7} & \multicolumn{1}{c|}{} & Accuracy ($\uparrow$) & Accuracy ($\uparrow$) & Accuracy ($\uparrow$) & Accuracy ($\uparrow$) & Accuracy ($\uparrow$) \\
\midrule
\multirow{1}{*}{\textbf{No Tuning}}
& Zero-shot (ZS)        & 59.44     & 52.79     & 11.82     & 43.48     & 38.61 \\
\midrule
\multirow{8}{*}{\textbf{ImageNet}}
& Linear-probing (LP)   & 72.43     & 61.35     & 10.71     & 41.58     & 38.19 \\
& Full Fine-tuning (FFT) & 73.14    & 60.98     & 6.41      & 32.71     & 32.83 \\[1ex]
\cline{2-7} \rule{0pt}{3ex}
& LP-FT \cite{kumar2022} & 73.30     & \textbf{62.04}     & 11.38      & 43.30     & 39.10 \\
& L2-SP \cite{Li2018LPSP} & 72.79     & 61.13     & 9.21      & 37.24     & 35.29 \\
& WiSE-FT \cite{wortsman2021} & \textbf{73.86}     & 61.50     & 11.27      & 42.18     & 37.92 \\
& CoOp \cite{zhou2022coop} & 70.64     & 58.12     & \textbf{13.89}      & 43.26     & 39.28 \\
\cline{2-7} \rule{0pt}{3ex}
& Reprogrammer (RP)     & 72.02$^{\pm 0.1}$     & 61.15$^{\pm 0.2}$     & 12.61$^{\pm 0.3}$     & 44.18$^{\pm 0.2}$     & 39.64$^{\pm 0.3}$ \\
& Residual Reprogrammer (RRP)     & 72.63$^{\pm 0.1}$     & 61.74$^{\pm 0.1}$     & 13.06$^{\pm 0.3}$     & \textbf{45.38$^{\pm 0.2}$}     & \textbf{40.12$^{\pm 0.2}$} \\
\bottomrule
\end{tabular}
}
\caption{\textbf{ImageNet Generalization Results.} OOD generalization performance comparison between zero-shot, linear-probing, full fine-tuning, L2-SP, WiSE-FT, CoOp, \name{}, and \resname{} methods. All methods utilize the CLIP B/32 architecture fine-tuned on \texttt{ImageNet-1k} as the in-distribution dataset.}
\label{tab:imagenet_generalization}
\vspace{-0.5cm}
\end{table*}

\subsection{Results}
\label{sec:results}

\begin{wraptable}{r}{0.5\textwidth}
\vspace{0cm}
    \centering
    {\footnotesize{
    \setlength\tabcolsep{2.15pt}
    \begin{tabular}{c|c|c|c}
    \toprule
    \multicolumn{1}{c|}{\multirow{2}{*}{}} & \multirow{2}{*}{\textbf{\begin{tabular}[c]{@{}c@{}}Method \end{tabular}}} & \scalebox{0.75}{\textbf{CIFAR Benchmark}} & \scalebox{0.75}{\textbf{ImageNet Benchmark}} \\
    \cline{3-4} & {} & \multicolumn{1}{c|}{{\scriptsize Aggregate} ($\uparrow$)} & \multicolumn{1}{c}{{\scriptsize Aggregate} ($\uparrow$)}\\
    \midrule
    \multirow{1}{*}{\textbf{No Tuning}}
    & ZS   & 88.22   & 51.54 \\
    \midrule
    \multirow{8}{*}{\textbf{Fine-tuned}}
    & LP   & 89.80   & 55.13 \\
    & FFT  & 81.66   & 51.88 \\
    [0.5ex] \cline{2-4} \rule{0pt}{3ex}
    & LP-FT   & 89.85   & 55.06 \\
    & L2-SP   & 87.49   & 53.39 \\
    & WiSE-FT & 89.58   & 55.75 \\
    & CoOp & 83.13   & 54.35 \\
    [0.5ex] \cline{2-4} \rule{0pt}{3ex}
    & RP   & 91.44   & 55.11 \\
    & RRP  & \textbf{92.69}   & \textbf{56.64} \\
    \bottomrule
    \end{tabular}}}
    \caption{\textbf{Aggregate Results} of the fine-tuned downstream model's performance across ID classification, OOD generalization, and OOD detection tasks. To quantify the holistic performance, we report the average score normalized across all benchmarks as described in the Experimental Setup in Section~\ref{sec:exp_setup}.}
    \label{tab:holistic_performance}
\vspace{-0.38cm}
\end{wraptable}

\vspace{-0.2cm}
\paragraph{Holistic Performance:} 
We present a holistic evaluation in Table~\ref{tab:holistic_performance}, showcasing an aggregated score based on the average normalized performance across ID, OOD generalization, and OOD detection tasks for \name{}, \resname{}, and other common fine-tuning techniques. More specifically, these aggregated scores are presented in relation to the specified in-distribution dataset (\texttt{CIFAR-10} or \texttt{ImageNet-1k}) utilized for fine-tuning the pre-trained model. We observe that the base \name{} method generally outperforms all other compared fine-tuning methods in terms of its aggregated performance. Furthermore, it is evident that the \resname{} method surpasses even \name{}, enhancing the holistic aggregate score by $+1.01\%$ in our \texttt{CIFAR-10} benchmarks and $+1.13\%$ in our \texttt{ImageNet-1k} benchmarks. These aggregated performances provide strong support for our hypotheses that less intrusive fine-tuning techniques, such as \name{}, will yield more holistically robust downstream models that are better equipped to handle covariate and semantically shifted OOD samples.

\begin{wraptable}{r}{0.5\textwidth}
\vspace{-0.45cm}
    \centering
    {\footnotesize{
    \setlength\tabcolsep{2.15pt}
    \begin{tabular}{c|c|c|c c}
    \toprule
    \multicolumn{1}{c|}{\multirow{2}{*}{\textbf{$\bm{D_{in}}$}}} & \multirow{2}{*}{\textbf{\begin{tabular}[c]{@{}c@{}}Method \end{tabular}}} & \textbf{CIFAR-10} & \textbf{CIFAR10.1} & \textbf{STL10} \\
    \cline{3-5} & {} & \multicolumn{1}{c|}{{\scriptsize Accuracy} ($\uparrow$)} & \multicolumn{1}{c}{{\scriptsize Accuracy} ($\uparrow$)} & \multicolumn{1}{c}{{\scriptsize Accuracy} ($\uparrow$)} \\
    \midrule
    \multirow{1}{*}{\textbf{No Tuning}}
    & ZS       & 89.23   & 83.30   & 97.40 \\
    \midrule
    \multirow{8}{*}{\textbf{CIFAR-10}}
    & LP  & 94.89   & 90.05   & 96.34 \\
    & FFT     & 96.24   & 91.05   & 55.90 \\
    [0.5ex] \cline{2-5} \rule{0pt}{3ex}
    & LP-FT & 96.38   & 91.53   & 95.93 \\
    & L2-SP & 95.46   & 90.71   & 87.59 \\
    & WiSE-FT & \textbf{97.63}   & 92.65   & 91.27 \\
    & CoOp & 94.50 & 90.45 & 68.94 \\
    [1ex] \cline{2-5} \rule{0pt}{3ex}
    & RP   & 95.23$^{\pm 0.1}$   & 91.42$^{\pm 0.1}$   & 96.58$^{\pm 0.3}$ \\
    & RRP  & 95.56$^{\pm 0.1}$   & \textbf{92.67$^{\pm 0.1}$}   & \textbf{97.86$^{\pm 0.1}$} \\
    \bottomrule
    \end{tabular}}
    }
    \caption{\textbf{CIFAR Generalization Results} OOD generalization performance comparison between zero-shot (ZS), linear-probing (LP), full fine-tuning (FFT), LP-FT, L2-SP, WiSE-FT, CoOp, \name{} (RP), and \resname{} (RRP) methods with \texttt{CIFAR-10} as the in-distribution dataset. RP and RRP results are averaged over 3 random seeds.}
    \label{tab:cifar_generalization}
\vspace{-0.38cm}
\end{wraptable}

\vspace{-0.2cm}
\paragraph{Out-of-distribution Generalization:} 
We provide a detailed evaluation in Table~\ref{tab:cifar_generalization} and Table~\ref{tab:imagenet_generalization}, focusing on the generalization accuracy of our \name{} and \resname{} methods after fine-tuning on the \texttt{CIFAR-10} and \texttt{ImageNet-1k} ID datasets, respectively. We note that WiSE-FT exceeds most other methods specifically in the ID classification task. This is in line with expectations set by prior works \cite{wortsman2021}. However, in the context of OOD generalization tasks, \resname{} consistently outperforms all other fine-tuning techniques across all benchmarks. This observation substantiates our hypothesis that maintaining diverse pre-trained representations is crucial for effectively generalizing to covariate-shifted OOD samples. Furthermore, we notice that full fine-tuning in particular yields significantly worse results compared to all other fine-tuning techniques. This observation also aligns with prior works that have shown naive full fine-tuning to negatively distort the pre-training representations necessary for robust OOD generalization \cite{kumar2022}.

\begin{table*}[t]
\centering
\setlength\tabcolsep{2.5pt}
\renewcommand{\arraystretch}{1.25}
\scriptsize{
\begin{tabular}{c|c|cc|cc|cc|cc|cc}
\toprule
\multicolumn{1}{c|}{\multirow{2}{*}{\textbf{$\bm{D_{in}}$}}} & \multicolumn{1}{c|}{\multirow{2}{*}{\textbf{Method}}} & \multicolumn{2}{c|}{\textbf{iSUN}} & \multicolumn{2}{c|}{\textbf{LSUN Resize}} & \multicolumn{2}{c|}{\textbf{Places365}} & \multicolumn{2}{c|}{\textbf{Textures}} & \multicolumn{2}{c}{\textbf{Average}} \\ 
\cline{3-12} \multicolumn{1}{c|}{} & \multicolumn{1}{c|}{} &  FPR95 ($\downarrow$) & AUROC ($\uparrow$) & FPR95 ($\downarrow$) & AUROC ($\uparrow$) & FPR95 ($\downarrow$) & AUROC ($\uparrow$) & FPR95 ($\downarrow$) & AUROC ($\uparrow$) & FPR95 ($\downarrow$) & AUROC ($\uparrow$) \\
\midrule
\multirow{1}{*}{\textbf{No Tuning}}
& ZS & 27.15 & 95.08 & 24.41 & 95.61 & 15.87 & 97.12 & 32.36 & 92.60 & 24.95 & 95.10 \\
\midrule
\multirow{8}{*}{\textbf{CIFAR-10}}
& LP & 36.74 & 94.57 & 28.38 & 95.75 & 24.65 & 96.73 & 39.67 & 92.93 & 32.36 & 94.99 \\
& FFT & 45.47 & 92.78 & 42.95 & 93.41 & 40.92 & 94.06 & 44.85 & 92.30 & 42.89 & 93.40 \\
[1ex] \cline{2-12} \rule{0pt}{3ex}
& LP-FT & 37.24 & 94.69 & 36.80 & 95.23 & 28.77 & 95.85 & 40.43 & 93.06 & 35.81 & 94.71 \\
& L2-SP & 40.20 & 93.76 & 34.94 & 94.47 & 35.09 & 94.84 & 42.72 & 92.71 & 38.24 & 93.95 \\
& WiSE-FT & 37.93 & 94.22 & 31.23 & 95.10 & 34.73 & 95.06 & 40.36 & 93.15 & 36.06 & 94.38 \\
& CoOp & 35.38 & 94.48 & 30.53 & 95.37 & 57.77 & 86.72 & 44.72 & 93.52 & 42.10 & 92.52 \\
[1ex] \cline{2-12} \rule{0pt}{3ex}
& RP  & 29.86$^{\pm 0.7}$ & 95.36$^{\pm 0.5}$ & 26.31$^{\pm 0.6}$ & 95.88$^{\pm 0.4}$ & 15.95$^{\pm 0.5}$ & 97.60$^{\pm 0.3}$ & 30.68$^{\pm 0.8}$ & 93.65$^{\pm 0.5}$ & 25.70$^{\pm 0.7}$ & 95.62$^{\pm 0.4}$\\
& RRP & \textbf{24.87$^{\pm 0.6}$} & \textbf{96.19$^{\pm 0.4}$} & \textbf{20.52$^{\pm 0.6}$} & \textbf{97.12$^{\pm 0.3}$} & \textbf{15.22$^{\pm 0.5}$} & \textbf{97.86$^{\pm 0.2}$} & \textbf{26.37$^{\pm 0.6}$} & \textbf{94.87$^{\pm 0.5}$} & \textbf{21.75$^{\pm 0.6}$} & \textbf{96.51$^{\pm 0.4}$}\\
\midrule

\midrule
\multicolumn{1}{c|}{\multirow{2}{*}{\textbf{$\bm{D_{in}}$}}} & \multicolumn{1}{c|}{\multirow{2}{*}{\textbf{Method}}} & \multicolumn{2}{c|}{\textbf{iNaturalist}} & \multicolumn{2}{c|}{\textbf{SUN}} & \multicolumn{2}{c|}{\textbf{Places}} & \multicolumn{2}{c|}{\textbf{Textures}} & \multicolumn{2}{c}{\textbf{Average}} \\ 
\cline{3-12} \multicolumn{1}{c|}{} & \multicolumn{1}{c|}{} &  FPR95 ($\downarrow$) & AUROC ($\uparrow$) & FPR95 ($\downarrow$) & AUROC ($\uparrow$) & FPR95 ($\downarrow$) & AUROC ($\uparrow$) & FPR95 ($\downarrow$) & AUROC ($\uparrow$) & FPR95 ($\downarrow$) & AUROC ($\uparrow$) \\
\midrule
\multirow{1}{*}{\textbf{No Tuning}}
& ZS & 53.96 & 85.15 & \textbf{64.89} & \textbf{81.26} & \textbf{65.76} & \textbf{79.30} & 70.05 & 77.03 & \textbf{63.67} & \textbf{80.69} \\
\midrule
\multirow{8}{*}{\textbf{ImageNet}}
& LP & \textbf{51.15} & 88.25 & 78.68 & 74.58 & 76.42 & 75.15 & 70.25 & \textbf{78.71} & 69.12 & 79.17 \\
& FFT & 71.94 & 81.37 & 80.29 & 74.01 & 79.97 & 74.54 & 78.28 & 74.80 & 77.62 & 76.18 \\
[1ex] \cline{2-12} \rule{0pt}{3ex}
& LP-FT & 59.28 & 85.51 & 78.74 & 74.56 & 76.51 & 75.01 & 73.88 & 76.67 & 72.10 & 77.94 \\
& L2-SP & 64.74 & 85.01 & 77.84 & 74.50 & 77.59 & 74.93 & 75.98 & 75.05 & 74.04 & 77.37 \\
& WiSE-FT & 52.21 & \textbf{88.63} & 77.08 & 74.89 & 75.70 & 75.91 & 71.05 & 77.92 & 69.01 & 79.34 \\
& CoOp & 60.51 & 82.91 & 77.63 & 73.88 & 76.12 & 73.71 & \textbf{64.70} & 78.68 & 69.74 & 77.30 \\
[1ex] \cline{2-12} \rule{0pt}{3ex}
& RP  & 57.13$^{\pm 1.1}$ & 85.82$^{\pm 0.7}$ & 76.68$^{\pm 1.5}$ & 74.31$^{\pm 1.0}$ & 75.89$^{\pm 1.8}$ & 74.32$^{\pm 1.1}$ & 70.53$^{\pm 1.6}$ & 77.09$^{\pm 1.0}$ & 70.06$^{\pm 1.5}$ & 77.89$^{\pm 1.0}$ \\
& RRP  & 51.46$^{\pm 0.9}$ & 87.82$^{\pm 0.6}$ & 69.95$^{\pm 1.1}$ & 77.29$^{\pm 0.9}$ & 70.93$^{\pm 1.5}$ & 76.58$^{\pm 1.0}$ & 69.10$^{\pm 1.5}$ & 77.48$^{\pm 1.0}$ & 65.36$^{\pm 1.3}$ & 79.79$^{\pm 0.9}$ \\
\bottomrule
\end{tabular}
}
\caption{\textbf{OOD Detection Results.} OOD detection performance comparison between zero-shot (ZS), linear-probing (LP), full fine-tuning (FFT), L2-SP, WiSE-FT, CoOp, \name{} (RP), and \resname{} (RRP) using the \texttt{msp} \cite{hendrycks2016baseline} detector. All methods utilize the CLIP B/32 architecture fine-tuned on \texttt{CIFAR-10} or \texttt{ImageNet-1k} as the in-distribution dataset. $\uparrow$ indicates larger values are better, while $\downarrow$ indicates smaller values are better. All values are percentages and \textbf{bold} values are the superior results.}
\label{tab:ood_detection}
\vspace{-0.2cm}
\end{table*}

\paragraph{Out-of-distribution Detection:}
We provide a detailed evaluation of OOD detection in Table~\ref{tab:ood_detection}. Specifically, we present the OOD detection performances of our fine-tuned models across four semantically shifted OOD datasets, as well as the averaged performance across all four datasets in both the \texttt{CIFAR-10} and the \texttt{ImageNet-1k} ID settings. To ensure fair comparisons, we employ the commonly used baseline \texttt{msp} detector \cite{hendrycks2016baseline} across all experiments as a measure to assess the level of overconfidence exhibited by each downstream model when dealing with semantically shifted OOD samples. Firstly, it is very apparent that all non-reprogrammer fine-tuning techniques exhibit worse OOD detection performance in comparison to the zero-shot model. This observation reinforces our hypothesis that fine-tuning techniques have an adverse impact on the downstream model's ability to detect semantically shifted OOD samples. 

Secondly, we also note that \resname{} outperforms all other fine-tuning techniques in both \texttt{ImageNet-1k} and \texttt{CIFAR-10} OOD detection benchmarks. However, in the \texttt{ImageNet-1k} benchmarks, neither \name{} nor \resname{} manages to surpass the OOD detection capabilities of the zero-shot model. The superiority of the zero-shot model for OOD detection, when compared to fine-tuned models is expected as recent research has also demonstrated the effectiveness of the zero-shot CLIP model for OOD detection tasks \cite{ming2022delving}. This implies that while \name{} and \resname{} lead to improved downstream models compared to other fine-tuning techniques, there remains an inherent OOD detection cost associated with fine-tuning a pre-trained model. Subsequently, this hidden cost can further materialize as a trade-off between generalization capabilities (ID \& OOD generalization) and detection capabilities (OOD detection) when fine-tuning a pre-trained model.

\subsection{Ablation Studies}
\label{sec:ablations}

\begin{figure*}[t]
    \captionsetup[subfigure]{justification=centering}
    \centering
    \begin{subfigure}{0.245\textwidth}
        \centering
        \includegraphics[width=\textwidth]{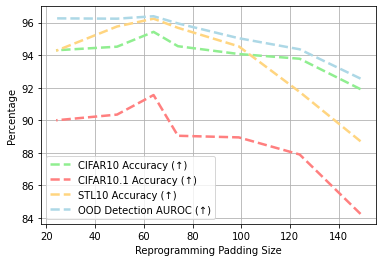}
        \caption{Reprogrammer Padding Size (CIFAR Benchmarks)}
        \label{fig:cifar_padding_ablation}
    \end{subfigure}
    \hfill
    \begin{subfigure}{0.245\textwidth}
        \centering
        \includegraphics[width=\textwidth]{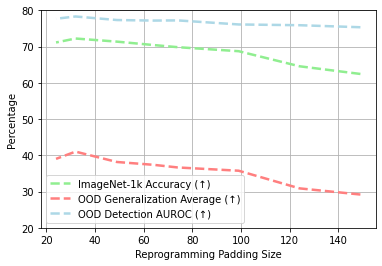}
        \caption{Reprogrammer Padding Size (ImageNet Benchmarks)}
        \label{fig:imagenet_padding_ablation}
    \end{subfigure}
    \begin{subfigure}{0.245\textwidth}
        \centering
        \includegraphics[width=\textwidth]{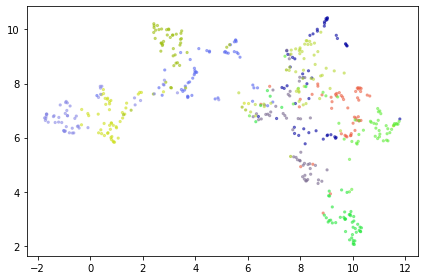}
        \caption{Linear-probing Feature Space}
        \label{fig:lp_embedding_ablation}
    \end{subfigure}
    \hfill
    \begin{subfigure}{0.245\textwidth}
        \centering
        \includegraphics[width=\textwidth]{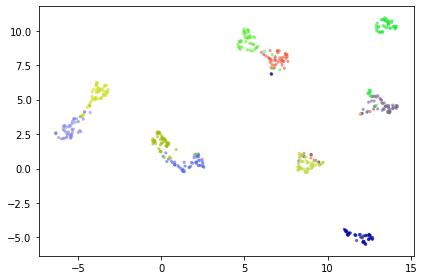}
        \caption{Reprogrammer Feature Space}
        \label{fig:rp_embedding_ablation}
    \end{subfigure}
\caption{\textbf{Ablation Studies.} Figures~\ref{fig:cifar_padding_ablation}, \ref{fig:imagenet_padding_ablation} illustrate the effectiveness of our \name{} method as we adjust the image reprogramming padding size. A larger padding size indicates that more of the input image is being subjected to the reprogramming function. Additionally, we present UMAP visualization comparing the feature spaces between linear-probed and \name{} models using $500$ randomly sampled covariate shifted (\texttt{CIFAR-10.1}) images in Figures~\ref{fig:lp_embedding_ablation}, \ref{fig:rp_embedding_ablation}.}
\label{fig:padding_ablation}
\vspace{-0.2cm}
\end{figure*}

\paragraph{Reprogrammer Padding Size:} 
In this ablation study, we evaluate the effectiveness of our \name{} training as we adjust the image reprogramming padding size. The image reprogramming padding size refers to a set of hyperparameters within our image reprogramming module $\psi$ that control the extent of border and padding perturbations applied to the image. Consequently, a larger image reprogramming padding size results in more extensive perturbations being applied through the image reprogramming module. We illustrate the effects of varying reprogramming padding sizes, increasing the permissible border pixels from $30$ to $140$, in Figure~\ref{fig:padding_ablation}. More specifically, we present the impacts of padding size adjustments on both our \texttt{CIFAR-10} benchmarks (Figure~\ref{fig:cifar_padding_ablation}) and \texttt{ImageNet-1k} benchmarks (Figure~\ref{fig:imagenet_padding_ablation}). Comparing the results of our ablation study, we observe that the optimal range of padding sizes for our \texttt{CIFAR-10} reprogrammed model tends to be larger than that for our \texttt{ImageNet-1k} reprogrammed model. We hypothesize that this discrepancy is due to the lower-resolution images present in the \texttt{CIFAR-10} dataset, which compels our \name{} to adopt a more aggressive perturbation strategy to compensate for the lower-resolution samples.

\vspace{-0.1cm}
\paragraph{Reprogrammed Feature Space:} 
In this ablation study, we offer additional insights to illustrate how reprogramming can enhance the alignment of covariate-shifted OOD samples. Figures~\ref{fig:lp_embedding_ablation} and \ref{fig:rp_embedding_ablation} present \textit{UMAP} visualizations comparing the feature space between the linear-probed and reprogrammed models on covariate-shifted OOD samples \cite{mcinnes2018}. Observing these visualizations, it becomes evident that \name{} generates more class-conditionally distinct and compact clusters of covariate features. This observation further substantiates our hypothesis that \textit{model reprogramming} techniques can effectively align OOD samples with the strongly tuned in-distribution space, thereby enabling \name{} and \resname{} to more accurately distinguish and classify covariate-shifted OOD samples.

\section{Discussion}
\label{sec:discussion}
In this section, we discuss the limitations and computational efficiency of our \name{} and \resname{} methods. Specifically, we explore some of the inherent limitations relating to Text-Image Encoder architectures in Section~\ref{sec:tiencoders}, discuss the challenges associated with learning under high output space dimensionality in Section~\ref{sec:dimensionality}, and elaborate on the computational efficiency of our methods in Section~\ref{sec:computational_efficiency}. Additionally, we also discuss our work in relation to some recent advancements in Appendix~\ref{app:comparison_maple}.

\vspace{-0.1cm}
\subsection{Text-Image Encoders}
\label{sec:tiencoders}
Our proposed \name{} and \resname{} methods aim to reprogram both the image and text encoders in CLIP-like models. Due in part to the necessity for there to be paired image and text encoders, our methods are limited to these joint text-image encoder models. However, diverging from our \name{} and \resname{} methods, we can harness components from our approach in combination with other fine-tuning techniques. Specifically, Kumar~\cite{kumar2022} demonstrated how to mitigate full fine-tuning distortions by initially tuning the classification head before proceeding with full fine-tuning of the model. Likewise, the image reprogramming function can be integrated with linear-probing or full fine-tuning to develop another set of fine-tuning techniques. These supplementary methods could potentially yield similar out-of-distribution benefits to those observed in our paper and may also be applicable to other non-CLIP pre-trained models. We leave this area open for future exploration.

\subsection{Output Space Dimensionality}
\label{sec:dimensionality}
A potential limitation of \name{} and \resname{} is their effectiveness in tasks with higher output dimensionality. Specifically, \textit{model reprogramming} techniques generally perform better on tasks with lower output dimensionality, such as the 10-way classification in \texttt{CIFAR-10}. However, when dealing with tasks with higher output dimensionality, like the 1000-way classification in \texttt{ImageNet-1k}, \textit{model reprogramming} techniques typically require more extensive tuning and tend to be less capable of outperforming other fine-tuning techniques. This limitation is reflected in our methods as, although \name{} and \resname{} still demonstrate improvements in the showcased ImageNet benchmarks, these improvements are less significant when compared to the CIFAR benchmarks. An intuitive initial approach to addressing this issue would be to replace our traditional and simple reprogramming functions with more robust functions. However, this topic is beyond the scope of this paper, and we will leave this problem open for future \textit{model reprogramming} research.

\subsection{Reprogramming Computational Efficiency}
\label{sec:computational_efficiency}
One of the key advantages of using \textit{model reprogramming} techniques lies in their minimal resource and data requirements, as demonstrated by Tsai et al \cite{tsai2020}. Specifically, the computational overhead introduced by the incorporation of \name{} is minimal. The image reprogramming function can be broken down into a masking function involving matrix addition, and the text reprogramming function is a simple lookup operation with vector addition. As a result, the training process incurs negligible overhead due to the small set of parameters needed to be learned. Furthermore, the memory complexity associated with maintaining the reprogramming functions is also minimal. It only necessitates the storage of matrices proportional to the fixed padding and vocabulary sizes specified by the \name{} methods. This efficiency in resource utilization makes \name{} and \resname{} an appealing choice for various real-world applications.

\section{Societal Impact}
\label{sec:societal}
The goal of our project is to enhance the safety and robustness of fine-tuning techniques applied to large pre-trained machine learning models. We believe that these improvements can have a profound impact across various societal domains. Given that many modern real-world applications heavily depend on classification, addressing these safety and robustness concerns is of paramount importance, spanning from consumer and business applications to autonomous vehicles and medical imaging. Through this endeavor, we aim to provide researchers with an additional tool to address these complex challenges. While we do not foresee any adverse consequences stemming from our work, we aspire to continue monitoring and building upon this method in the future.

\section{Conclusion}
\label{sec:conclusion}
In this paper, we demonstrated that preserving pre-training representations is critical for improving the holistic capabilities (ID classification, robustness to covariate shifts, and safety under semantic shifts) of the downstream model. To this end, we introduced an alternative approach for fine-tuning text-image encoder models called \name{}, which aims to minimize distortion to the model's pre-trained representations through \textit{model reprogramming} techniques. Experimental results further highlight the effectiveness of both \name{} and \resname{} when compared to other common fine-tuning techniques. We hope that our study illuminates the hidden costs associated with common fine-tuning techniques and inspires future research to leverage reprogramming approaches for fine-tuning. Moreover, we hope that our study helps to underscore the importance of measuring holistic ID and OOD performances (Table~\ref{tab:holistic_performance}) when evaluating the effectiveness of different fine-tuning techniques.

\bibliography{arxiv}
\bibliographystyle{plainnat}

\newpage
\appendix
\onecolumn

\begin{center}
    \Large{\textbf{Supplementary Material}}
\end{center}

\section{Details of Experiments}
\label{app:exp_detail}
In this Appendix, we present a detailed description of the chosen OOD generalization and OOD detection datasets in Appendix~\ref{app:datasets} and a description of our software and hardware specifications in Appendix~\ref{app:software_hardware}.

\subsection{Datasets}
\label{app:datasets}
We present a detailed list of our OOD generalization and OOD detection evaluation datasets, along with a brief description of each dataset.

\paragraph{CIFAR-10 OOD Generalization Benchmarks:}
\begin{itemize}
    \item \texttt{CIFAR-10.1}~\cite{torralba2008} is a collection of over 2,000 test images, sampled from TinyImages, which are designed to be a minute distributional shift from the CIFAR-10 dataset.
    \item \texttt{STL10}~\cite{coates2011} is a collection of over 8,000 test images, sampled from \texttt{ImageNet-1k}, that is commonly used in domain adaptation studies. We carefully curate the STL10 dataset to evaluate with only the 9 semantically overlapping classes, choosing to omit the semantically different "monkey" class. 
\end{itemize}

\paragraph{ImageNet-1k OOD Generalization Benchmarks:}
\begin{itemize}
    \item \texttt{ImageNetV2}~\cite{recht2019} is a collection of 10,000 test images with approximately 10 samples per class. The dataset was sampled utilizing the same semantic labels as defined in \texttt{ImageNet-1k} and obtained independently from any previous ImageNet models.
    \item \texttt{ImageNet-A}~\cite{hendrycks2021nae} is a collection of 7,500 naturally adversarial and challenging images that are sampled based on 200 semantically overlapping \texttt{ImageNet-1k} classes.
    \item \texttt{ImageNet-R}~\cite{hendrycks2021many} is a collection of over 30,000 test images, based on 200 semantically overlapping \texttt{ImageNet-1k} classes, that contain images of art, cartoon, graffiti, embroidery, origami, toy, sculpture, sketch, tattoo, and other rendition of the \texttt{ImageNet-1k} classes.
    \item \texttt{ImageNet-Sketch}~\cite{wang2019learning} is a collection of over 50,000 test images based on all 1000 of the \texttt{ImageNet-1k} classes with approximately 50 images per class. Each image is a black-and-white sketch variant of the \texttt{ImageNet-1k} class.
\end{itemize}

\paragraph{CIFAR-10 OOD Detection Benchmarks:}
\begin{itemize}
    \item \texttt{iSUN}~\cite{pingmei2015} is a collection of over 8,925 natural scene images sampled from the SUN dataset. We include the full set of iSUN images when conducting OOD detection evaluations.
    \item \texttt{LSUN Resized}~\cite{yu15lsun} is a collection of 10,000 testing images, sampled from the LSUN dataset, spanning across 10 different scenes with images down-sampled to the size of (32$\times$32). We include the full set of LSUN Resized images when conducting OOD detection evaluations.
    \item \texttt{Places365}~\cite{zhou2017places} contains large-scale photographs of scenes with 365 scene categories. There are 900 images per category in the test set and we again include the full test set for OOD detection evaluations.
    \item \texttt{Textures}~\cite{cimpoi14describing}, or Describable Textures Dataset, is a collection of 5,640 real-world texture images under 47 categories. We include the entire set of 5640 images for OOD detection evaluations.
\end{itemize}

\paragraph{ImageNet-1k OOD Detection Benchmarks:}
\begin{itemize}
    \item \texttt{iNaturalist}~\cite{van2018inaturalist} is a collection of 859,000 plant and animal images spanning over 5,000 different species. Each image is resized to have a max dimension of 800 pixels and we evaluate 10,000 images randomly sampled from 110 classes that are carefully chosen to be semantically disjoint from the \texttt{ImageNet-1k} dataset.
    \item \texttt{SUN}~\cite{xiao2010sun} is a collection of over 130,000 images of scenes spanning 397 categories. We evaluate 10,000 randomly sampled images from 50 classes that are semantically disjoint from \texttt{ImageNet-1k} classes, as \texttt{SUN} and \texttt{ImageNet-1k} have overlapping semantic concepts.
    \item \texttt{Places}~\cite{zhou2017places} is a collection of scene images with similar semantic coverage as SUN. We use a subset of 10,000 images across 50 classes that are semantically disjoint from the \texttt{ImageNet-1k} dataset.
    \item \texttt{Textures}~\cite{cimpoi14describing}, or Describable Textures Dataset, is a collection of 5,640 real-world texture images under 47 categories. We again include the entire set of 5640 images for OOD detection evaluations.
\end{itemize}

\subsection{Software and Hardware}
\label{app:software_hardware}
\paragraph{Software} We conducted all experiments with Python 3.8.12 and PyTorch 1.11.0.

\paragraph{Hardware} All experiments were conducted on NVIDIA GeForce RTX 2080Ti.

\subsection{Evaluation Metrics} 
\label{app:evaluation_metrics}
In the context of OOD generalization, we measure all methods across the designated covariate-shifted OOD datasets using accuracy as the evaluation metric. For OOD detection, we measure all methods across each semantically shifted dataset using the false positive rate, when the true positive rate of ID samples is 95\% (FPR95), and the area under the receiver operating characteristic curve (AUROC) as evaluation metrics.

\subsection{Learning Details}
\label{app:learning_details}
All presented experiments were conducted using the CLIP B/32 architecture unless otherwise specified. Additional experiments with higher capacity models can be found in Appendix~\ref{app:additional_experiments}. We conducted a simple hyperparameter sweep for \resname{}, varying $\alpha$ from $\{0.0,0.1,\ldots, 1.0\}$, based on ID test accuracy. The final chosen value used during the evaluation was $\alpha=0.4$. For images with a resolution higher than $128\times 128$, we downscaled and cropped them to $128\times 128$. We did this to accommodate CLIP's input size limitations, ensure paddable pixels, and maintain a fair comparison across all datasets, considering the information loss due to downsampling. Additional experiments with various degrees of downsampling, ranging from no downsampling to heavy downsampling, are presented in Appendix~\ref{app:down_sampling}. In all fine-tuning training processes, we initialized the model with the pre-trained CLIP B/32 model and performed a hyperparameter sweep over three learning rates using a cosine learning rate scheduler. For linear probing, we directly optimized a linear regression classifier on the frozen features extracted from the penultimate layer of the CLIP image encoder, sweeping over learning rates of $\{0.005, 0.002, 0.001\}$ for 5 epochs. For full fine-tuning, we initialized the classification head with text encoder features derived from class-wise captions, as specified by Wortsman~\cite{wortsman2021}. We then conducted a sweep over learning rates of $\{0.00001, 0.00003, 0.0001\}$ for 5 epochs, optimizing all parameters in the image encoder and classification head. For LP-FT, we initialize the classification head using linear probing before full fine-tuning the model. Additionally, for WiSE-FT, we utilized $\alpha=0.5$. For \name{}, we randomly initialized both the image and text reprogramming functions and conducted a sweep over learning rates of $\{0.0005, 0.001, 0.005\}$ for 5 epochs. In all experiments, we set the batch size to 128 and included a warm-up period of 500 iterations. Further details regarding hyperparameter settings can be found in the provided source code. In addition, when tuning on \texttt{CIFAR-10} and \texttt{ImageNet-1k}, we set the image up-sampling for \name{} and \resname{} to $160\times 160$ and $224\times 224$ pixels, respectively, with padding sizes of 64 and 32. As previously mentioned, images with resolutions higher than $128\times 128$ were downscaled and cropped to $128\times 128$ to accommodate CLIP's input size limitations and ensure consistent comparisons across datasets. Additional experiments featuring varying degrees of downsampling, from none to heavy downsampling, are presented in the Appendix~\ref{app:down_sampling}.

\subsection{Compared Methods}
\label{app:compared_methods}
We compared our \name{} (RP) and \resname{} (RRP) fine-tuned models against zero-shot (ZS), linear-probed (LP), full fine-tuned (FFT), LP-FT \cite{kumar2022}, L2-SP \cite{Li2018LPSP}, WiSE-FT \cite{wortsman2021}, and CoOp \cite{zhou2022coop} fine-tuned models. Each of these fine-tuning methods is commonly used in CLIP-based OOD evaluations. Zero-shot refers to applying the CLIP pre-trained model directly to the designated downstream task without making any alterations to the CLIP model. Linear probing involves optimizing a linear regression classifier directly on the frozen features extracted from the penultimate layer of the CLIP image encoder. To obtain a fully fine-tuned model, we fine-tuned all parameters in the image encoder and classification head to fit the in-distribution dataset. Subsequently, both L2-SP and WiSE-FT can be considered more sophisticated alternatives to full fine-tuning. Specifically, L2-SP is a regularized full fine-tuning method that applies L2 regularization to the model's parameters with respect to the zero-shot model parameters \cite{Li2018LPSP}. WiSE-FT is a model souping approach that combines the weights of a fully fine-tuned model with the zero-shot model \cite{wortsman2021}. Finally, CoOp is a prompt learning approach to fine-tuning where the learned parameters consist solely of class-wise captions as defined by $s_i$ \cite{zhou2022coop}.

\section{Details of Methodology}
\label{app:detail_method}
In this appendix, we present additional visualizations to help explain the individual components of \name{} in detail. We first present additional image reprogramming details and visualizations, before moving on to detailing the text reprogramming component.

\subsection{Image Reprogramming}
\label{app:image_reprogramming}
We present a visual diagram of our image reprogramming in the top half of Figure~\ref{app:reprogrammer_methodology}. Additionally, we note that the image reprogramming can up-sample images to the input size of the pre-trained model. However, due to restrictions in the current open-sourced pre-trained CLIP models, our image reprogramming up-sampling is limited to being $3\times 224\times 224$ dimensions or less. Additionally, as part of the reprogramming function $\psi$, the size of the up-sampling/padding function $\mathcal{U}$ and binary masking matrix $M$ are tunable hyperparameters.

\subsection{Text Reprogramming}
\label{app:text_reprogramming}
Similarly, we present a visual diagram of our text reprogramming function in the bottom half of Figure~\ref{app:reprogrammer_methodology}. We generate class-wise captions following closely with the experiments presented by Radford et al \cite{radford2021}. Specifically, we set $s_i = ``\text{a photo of a }\{c_i\} "$ where $c_i$ is the given sample class label. As an example, our text reprogramming follows the procedures where, given a text Labrador Retriever label, our text reprogramming first tokenizes the string $s = ``\text{a photo of a Labrador Retriever}"$ into tokens $t_s$. Subsequently, the tokens $t_s$ are passed into the $\Phi_\theta$ function to embed the tokens into a vector $v'_s$. Then we apply a bias parameter $b$ to the given vector $v'_s$ in the form of $v_s = v'_s + b$, before finally passing the vector $v_s$ through the CLIP text encoder $g$ to get the reprogrammed text features.

\section{Analyzing Representation Residual Connection}
\label{app:analysis}
\noindent 
For ease of notation, let us consider
\begin{align*}
f(x) &= f(x)\:/\:\text{norm}(f(x))\\
g(s_i) &= g(s_i)\:/\:\text{norm}(g(s_i))\\
\hat{f}(x) &= f( \psi_{W^*}(x) )\:/\:\text{norm}(f( \psi_{W^*}(x) ))\\
\hat{g}(s_i) &= g(\Phi_{\theta^*, b^*}(s_i))\:/\:\text{norm}(g(\Phi_{\theta^*, b^*}(s_i)))
\end{align*}
where $\text{norm}(\cdot)$ represents the vector $l_2$ norm.

\noindent
Therefore, we can reduce our \resname{} classification to:
\begin{align}
y &= \argmax_i\left[ \similarity \left( (1-\alpha)\hat{f}(x) + \alpha f(x), ((1-\alpha)\hat{g}(s_i) + \alpha g(s_i) \right) \right]\\
&= \argmax_i[ ( (1-\alpha)\hat{f}(x) + \alpha f(x) )^\top ((1-\alpha)\hat{g}(s_i) + \alpha g(s_i) ) ]\\
&= \argmax_i[ (1-\alpha)^2\hat{f}(x) \hat{g}(s_i) + \alpha(1-\alpha) \hat{f}(x)g(s_i) + \alpha(1-\alpha) f(x)\hat{g}(s_i) + \alpha^2 f(x) g(s_i)]\\
&= \argmax_i[ (1-\alpha)^2\hat{f}(x) \hat{g}(s_i) + \alpha(1-\alpha) (\hat{g}(s_i)+\epsilon)g(s_i) + \alpha(1-\alpha) f(x)(\hat{f}(x)-\epsilon) + \alpha^2 f(x) g(s_i)]\\
&= \argmax_i[ (1-\alpha)^2 \hat{f}(x) \hat{g}(s_i)  + \alpha^2 f(x) g(s_i) + \alpha(1-\alpha) \hat{g}(s_i) g(s_i)\notag\\
&\qquad \qquad \;\; + \alpha(1-\alpha) f(x)\hat{f}(x) + \alpha(1-\alpha) (\epsilon g(s_i) - \epsilon f(x)) ]\\
&= \argmax_i[ (1-\alpha)^2 \hat{f}(x) \hat{g}(s_i)  + \alpha^2 f(x) g(s_i) + \alpha(1-\alpha) \hat{g}(s_i) g(s_i)\notag\\
&\qquad \qquad \;\; + \alpha(1-\alpha) f(x)\hat{f}(x) + \alpha(1-\alpha) (\hat{f}(x) - \hat{g}(s_i)) (g(s_i) - f(x)) ]
\end{align}
where (10) holds given $\epsilon = \hat{f}(x) - \hat{g}(s_i)$. Subsequently, we can now see that our \resname{} is simply a complex combination of the \name{} classification $(1-\alpha)^2 \hat{f}(x) \hat{g}(s_i)$, zero-shot classification $\alpha^2 f(x) g(s_i)$, some regularization by the zero-shot representation $\alpha(1-\alpha) \hat{g}(s_i) g(s_i)$, and some additional closeness regularization $\alpha(1-\alpha) (\hat{f}(x) - \hat{g}(s_i)) (g(s_i) - f(x))$.

\section{Additional Discussion}
\label{app:discussion}
In this appendix, we present some additional discussion regarding the use of differing architectures in Appendix~\ref{app:differing_architectures}.

\begin{table*}[t]
\centering
\setlength\tabcolsep{7pt} 
\renewcommand{\arraystretch}{1.1}
\scriptsize{
\begin{tabular}{c|c|c|c|c|c|c}
\toprule
\multicolumn{1}{c|}{\multirow{3}{*}{\textbf{$\bm{D_{in}}$}}} & \multirow{3}{*}{\textbf{\begin{tabular}[c]{@{}c@{}} Method \end{tabular}}} & \multicolumn{1}{c|}{\textbf{ImageNet-1k}} & \multicolumn{1}{c|}{\textbf{ImageNetV2}} & \multicolumn{1}{c|}{\textbf{ImageNet-A}} & \multicolumn{1}{c|}{\textbf{ImageNet-R}} & \multicolumn{1}{c}{\textbf{ImageNet-S}} \\ 
\cline{3-7} & \multicolumn{1}{c|}{} & Accuracy & Accuracy & Accuracy & Accuracy & Accuracy \\
& \multicolumn{1}{c|}{} & \multicolumn{1}{c|}{$\uparrow$} & \multicolumn{1}{c|}{$\uparrow$} & \multicolumn{1}{c|}{$\uparrow$} & \multicolumn{1}{c|}{$\uparrow$} & \multicolumn{1}{c}{$\uparrow$}\\ 
\midrule
\multirow{1}{*}{\textbf{No Tuning}}
& Zero-shot (ZS)        & 75.26    & 64.13     & 27.98     & 52.42     & 39.80 \\
\midrule
\multirow{4}{*}{\textbf{ImageNet}}
& Linear-probing (LP)   & \textbf{83.90}    & 69.72     & 46.13     & 68.71     & 41.58 \\
[1ex] \cline{2-7} \rule{0pt}{3ex}
& Reprogrammer (RP)     & 83.42    & 70.36     & 51.62     & 70.20     & 43.80 \\
& Residual Reprogrammer (RRP)     & 83.61    & \textbf{70.92}     & \textbf{52.45}     & \textbf{72.97}     & \textbf{44.28} \\
\bottomrule
\end{tabular}
}
\caption{\textbf{CLIP L/14 ImageNet Generalization Results} OOD generalization performance comparison between zero-shot, linear-probing, \name{}, and \resname{} methods. All methods utilize the CLIP L/14 architecture fine-tuned on \texttt{ImageNet-1k} as the in-distribution dataset. The description of the four covariate shifted OOD datasets is provided in the Appendix. $\uparrow$ indicates larger values are better, while $\downarrow$ indicates smaller values are better. All values are percentages and \textbf{bold} values are the superior results.}
\vspace{-0.1cm}
\label{app:large_imagenet_generalization}
\end{table*}

\subsection{Differing Architectures}
\label{app:differing_architectures}
Within our evaluations, we leverage CLIP as the pre-trained model to which we apply our \name{} and \resname{} methods. Subsequently, a natural question arises asking how effective would our methods be when applied to other similar CLIP-like models with differing encoder architectures and training datasets such as ALIGN \cite{jia2021} and BASIC \cite{pham2021}. But, due in large part to a lack of available open-source pre-trained model parameters, we are unable to train or test with these comparable models. However, critically it is important to note that our \name{} methodology is not inherently limited in any way to just the open-source CLIP models. Specifically, ALIGN and BASIC primarily differ from CLIP only in their scale, as both ALIGN and BASIC can be interpreted as CLIP but with larger capacity transformer architectures alongside a larger training dataset. Therefore, as both ALIGN and BASIC are fundamentally similar to CLIP, we hypothesize that \name{} and \resname{} should show similar effectiveness when applied to either ALIGN or BASIC. Subsequently, we leave this question open for future exploration and encourage researchers, with more readily available resources, to experiment with our proposed methodologies.

\begin{table}[b]
\centering
\setlength\tabcolsep{5.15pt}
\begin{tabular}{c|c|c|c c}
\toprule
\multicolumn{1}{c|}{\multirow{2}{*}{\textbf{$\bm{D_{in}}$}}} & \multirow{2}{*}{\textbf{\begin{tabular}[c]{@{}c@{}}Method \end{tabular}}} & \textbf{CIFAR-10} & \textbf{CIFAR10.1} & \textbf{STL10} \\
\cline{3-5} & {} & \multicolumn{1}{c|}{{\scriptsize Accuracy} ($\uparrow$)} & \multicolumn{1}{c}{{\scriptsize Accuracy} ($\uparrow$)} & \multicolumn{1}{c}{{\scriptsize Accuracy} ($\uparrow$)} \\
\midrule
\multirow{1}{*}{\textbf{No Tuning}}
& ZS       & 96.18   & 82.67   & 99.53 \\
\midrule
\multirow{3}{*}{\textbf{CIFAR-10}}
& LP  & 98.04   & 94.63   & 86.29 \\
[1ex] \cline{2-5} \rule{0pt}{3ex}
& RP  & 98.32   & 95.72   & 98.79 \\
& RRP  & \textbf{98.64}   & \textbf{96.16}   & \textbf{99.86} \\
\bottomrule
\end{tabular}
\vspace{0.3cm}
\caption{\textbf{CLIP L/14 CIFAR Generalization Results} OOD generalization performance comparison between zero-shot (ZS), linear-probing (LP), and \name{} (RP) methods utilizing CLIP ViT-L/14 tuned with \texttt{CIFAR-10} as the in-distribution dataset. $\uparrow$ indicates larger values are better, while $\downarrow$ indicates smaller values are better. All values are percentages and \textbf{bold} values are the superior results.}
\label{app:large_cifar_generalization}
\end{table}

\section{Higher Capacity CLIP Experiments}
\label{app:additional_experiments}
In this appendix, we present additional experimental results showcasing the performance of \name{} and \resname{} when using higher capacity CLIP models. Specifically, we present OOD generalization performances and OOD detection performances with the larger CLIP L/14 model.

\vspace{-0.1cm}
\subsection{OOD Generalization}
\label{app:sec_ood_generalization}
We present the ImageNet OOD generalization results in Table~\ref{app:large_imagenet_generalization} and the CIFAR OOD generalization results in Table~\ref{app:large_cifar_generalization} using the large pre-trained CLIP L/14 model. We choose to omit full-fine-tuning experiments due to limited limited computational resources.
We observe that similar to our experimental observations with the B/32 CLIP model, \name{} and \resname{} consistently outperform both linear-probing and zero-shot on all of our OOD generalization benchmarks. 

\vspace{-0.1cm}
\subsection{OOD Detection}
\label{app:sec_ood_detection}
We present our OOD detection in Table~\ref{app:large_ood_detection} using the large pre-trained CLIP L/14 model. Specifically, in the top half of Table~\ref{app:large_ood_detection}, we report the OOD detection performance with the CIFAR benchmarks, and in the bottom half of Table~\ref{app:large_ood_detection} we report the OOD detection performance with the ImageNet benchmarks. Due to limited computational resources, we have chosen to omit full fine-tuning results. Again, for a fair comparison, we use the same commonly used baseline \texttt{msp} detector across all experiments as a way to gauge the level of overconfidence the zero-shot, linear-probed, \name{}, and \resname{} models has on semantically shifted OOD samples. 

We can see that similar to the CLIP B/32 experiments, our \resname{} outperforms all other fine-tuned models. However, again following observations with the CLIP B/32 experiments, we see that none of the fine-tuned downstream models were able to exceed the OOD detection capabilities of the zero-shot model. This again reaffirms the hypothesis that there is a hidden cost associated with fine-tuning a pre-trained model. In particular, this hidden cost seems to be most prominent when observing the capabilities of the downstream model on OOD detection tasks.

\begin{table*}[t]
\centering
\setlength\tabcolsep{5.15pt}
\renewcommand{\arraystretch}{1.25}
\scriptsize{
\begin{tabular}{c|c|cc|cc|cc|cc|cc}
\toprule
\multicolumn{1}{c|}{\multirow{3}{*}{\textbf{$\bm{D_{in}}$}}} & \multicolumn{1}{c|}{\multirow{3}{*}{\textbf{Method}}} & \multicolumn{2}{c|}{\textbf{iSUN}} & \multicolumn{2}{c|}{\textbf{LSUN Resize}} & \multicolumn{2}{c|}{\textbf{Places365}} & \multicolumn{2}{c|}{\textbf{Textures}} & \multicolumn{2}{c}{\textbf{Average}} \\ 
\cline{3-12} \multicolumn{1}{c|}{} & \multicolumn{1}{c|}{} &  FPR95 & AUROC & FPR95 & AUROC & FPR95 & AUROC & FPR95 & AUROC & FPR95 & AUROC \\
\multicolumn{1}{c|}{} & \multicolumn{1}{c|}{} & \multicolumn{1}{c}{$\downarrow$} & \multicolumn{1}{c|}{$\uparrow$} & \multicolumn{1}{c}{$\downarrow$} & \multicolumn{1}{c|}{$\uparrow$} & \multicolumn{1}{c}{$\downarrow$} & \multicolumn{1}{c|}{$\uparrow$} & \multicolumn{1}{c}{$\downarrow$} & \multicolumn{1}{c|}{$\uparrow$} & \multicolumn{1}{c}{$\downarrow$} & \multicolumn{1}{c}{$\uparrow$} \\
\midrule
\multirow{1}{*}{\textbf{No Tuning}}
& ZS & 11.65 & 97.51 & 17.07 & 96.39 & 10.92 & 97.60 & 27.58 & 93.09 & 16.81 & 96.15\\
\midrule
\multirow{3}{*}{\textbf{CIFAR-10}}
& LP  & 19.58 & 96.47 & 25.96 & 96.08 & 15.94 & 97.43 & 30.13 & 93.12 & 22.90 & 96.04\\
[1ex] \cline{2-12} \rule{0pt}{3ex}
& RP  & 10.31 & 97.20 & 15.84 & 96.53 & 12.08 & 97.63 & 20.51 & 94.75 & 14.68 & 96.53\\
& RRP & \textbf{7.10} & \textbf{98.24} & \textbf{11.90} & \textbf{97.95} & \textbf{10.57} & \textbf{98.01} & \textbf{16.81} & \textbf{97.06} & \textbf{11.60} & \textbf{97.81}\\
\midrule

\midrule
\multicolumn{1}{c|}{\multirow{3}{*}{\textbf{$\bm{D_{in}}$}}} & \multicolumn{1}{c|}{\multirow{3}{*}{\textbf{Method}}} & \multicolumn{2}{c|}{\textbf{iNaturalist}} & \multicolumn{2}{c|}{\textbf{SUN}} & \multicolumn{2}{c|}{\textbf{Places}} & \multicolumn{2}{c|}{\textbf{Textures}} & \multicolumn{2}{c}{\textbf{Average}} \\ 
\cline{3-12} \multicolumn{1}{c|}{} & \multicolumn{1}{c|}{} &  FPR95 & AUROC & FPR95 & AUROC & FPR95 & AUROC & FPR95 & AUROC & FPR95 & AUROC \\
\multicolumn{1}{c|}{} & \multicolumn{1}{c|}{} & \multicolumn{1}{c}{$\downarrow$} & \multicolumn{1}{c|}{$\uparrow$} & \multicolumn{1}{c}{$\downarrow$} & \multicolumn{1}{c|}{$\uparrow$} & \multicolumn{1}{c}{$\downarrow$} & \multicolumn{1}{c|}{$\uparrow$} & \multicolumn{1}{c}{$\downarrow$} & \multicolumn{1}{c|}{$\uparrow$} & \multicolumn{1}{c}{$\downarrow$} & \multicolumn{1}{c}{$\uparrow$} \\
\midrule
\multirow{1}{*}{\textbf{No Tuning}}
& ZS & \textbf{30.07} & \textbf{95.82} & \textbf{41.37} & \textbf{94.06} & \textbf{42.96} & \textbf{93.96} & \textbf{42.13} & \textbf{92.59} & \textbf{36.63} & \textbf{94.11} \\
\midrule
\multirow{3}{*}{\textbf{ImageNet}}
& LP & 42.58 & 93.72 & 51.49 & 88.12 & 56.98 & 88.73 & 58.95 & 87.89 & 52.50 & 89.62 \\
[1ex] \cline{2-12} \rule{0pt}{3ex}
& RP  & 45.79 & 93.31 & 49.37 & 90.03 & 54.93 & 88.90 & 59.36 & 87.73 & 52.36 & 89.99 \\
& RRP  & 40.46 & 94.18 & 42.81 & 91.55 & 50.62 & 90.75 & 50.20 & 90.06 & 46.02 & 91.64 \\
\bottomrule
\end{tabular}
}
\vspace{0.2cm}
\caption{\textbf{CLIP L/14 OOD Detection Results} OOD detection performance comparison between zero-shot (ZS), linear-probing (LP), \name{} (RP), and \resname{} (RRP) methods using the \texttt{msp} \cite{hendrycks2016baseline} detector. All methods utilize the CLIP L/14 architecture fine-tuned on \texttt{CIFAR-10} or \texttt{ImageNet-1k} as the in-distribution dataset. $\uparrow$ indicates larger values are better, while $\downarrow$ indicates smaller values are better. All values are percentages and \textbf{bold} values are the superior results.}
\vspace{-0.2cm}
\label{app:large_ood_detection}
\end{table*}

\section{Comparison with MaPLe}
\label{app:comparison_maple}
Khattak et al. proposed a new prompt learning technique called MaPLe, specifically designed for multi-modal models \cite{khattak2022MaPLe}. MaPLe was developed concurrently with \name{} and \resname{}, and both methodologies address the same domain of multi-modal model fine-tuning. However, there are significant methodological differences that distinguish \name{} and \resname{} from MaPLe. In particular, the Deep Vision Prompting in MaPLe utilizes multi-layer prompting, where each layer of the transformer undergoes an independent prompt learning module \cite{khattak2022MaPLe}. This approach differs from both traditional model reprogramming and RP, as they employ solely an input-level transformation function. Subsequently, this enables RP to be more lightweight, easier to implement in real-world applications, and versatile for settings like black-box optimization. Additionally, the Vision Language Prompt Coupling proposed in MaPLe establishes fixed prompting pairs between all layers of the Image and Text encoders \cite{khattak2022MaPLe}. This deviates from RP, where each modality is provided with its reprogramming function that is independently learned. This design proves to be particularly advantageous for diverse multi-modal models, where the paired modalities may not be text-image. For example, multi-modal models of text-audio will prove challenging for MaPLe to adapt to. 

We would like to reiterate that MaPLe is specifically designed to enhance prompt tuning for In-Distribution (ID) classification and generalization settings. In contrast, our work focuses on addressing robustness through Out-of-Distribution (OOD) detection and generalization. Furthermore, our goal is to shed light on the challenging trade-off between In-Distribution (ID) and OOD performances that can arise during the fine-tuning process. Additionally, MaPLe focuses on ID classification and generalization settings, while our work specifically addresses the OOD generalization and OOD detection settings. Consequently, a substantial portion of our paper aims to shed light on the challenging trade-off between ID and OOD performance that can arise during fine-tuning, and we also provide an in-depth discussion on how to measure and address these trade-off concerns. In summary, there are significant methodological and setting differences between our work and MaPLe, making direct comparisons between the methods challenging.

\begin{table}[t]
\centering
{\footnotesize{
\setlength\tabcolsep{5.15pt}
\begin{tabular}{c|c|c|c c|c}
\toprule
\multicolumn{1}{c|}{\multirow{2}{*}{\textbf{$\bm{D_{in}}$}}} & \multirow{2}{*}{\textbf{\begin{tabular}[c]{@{}c@{}}Method \end{tabular}}} & \textbf{CIFAR-10} & \textbf{CIFAR10.1} & \textbf{STL10} & \textbf{Aggregate} \\
\cline{3-6} & {} & \multicolumn{1}{c|}{{\scriptsize Accuracy} ($\uparrow$)} & \multicolumn{1}{c}{{\scriptsize Accuracy} ($\uparrow$)} & \multicolumn{1}{c|}{{\scriptsize Accuracy} ($\uparrow$)} & \multicolumn{1}{c}{{\scriptsize Aggregate} ($\uparrow$)} \\
\midrule
\multirow{3}{*}{\textbf{CIFAR-10}}
& OE       & \textbf{95.62}   & 90.12   & 68.49 & 89.79 \\
[1ex] \cline{2-6} \rule{0pt}{3ex}
& RP   & 95.23$^{\pm 0.1}$   & 91.42$^{\pm 0.1}$   & 96.58$^{\pm 0.3}$ & 91.44 \\
& RRP  & 95.56$^{\pm 0.1}$   & \textbf{92.67$^{\pm 0.1}$}   & \textbf{97.86$^{\pm 0.1}$} & \textbf{92.69} \\
\bottomrule
\end{tabular}}
}
\vspace{0.2cm}
\caption{\textbf{Results.} OOD generalization and aggregate performance comparison with outlier exposure (OE) using \texttt{CIFAR-10} as the in-distribution dataset. Values are percentages and \textbf{bold} values are the superior results.}
\label{tab:oe_cifar_generalization}
\end{table}

\section{Comparison with Outlier Exposure}
\label{app:comparison_oe}
\vspace{-0.2cm}
In this section, we provide a brief comparison between \name{} and \resname{} with full fine-tuned outlier exposure \cite{hendrycks2018deep}. We trained OE using the same full fine-tuning training setting as specified in Section~\ref{sec:exp_setup} alongside TinyImages~\cite{torralba200880} as the auxiliary OOD dataset and a $\beta =0.5$ as specified by Hendrycks et al. \cite{hendrycks2018deep}. Observing the aggregated results in Table~\ref{tab:oe_cifar_generalization}, we note that RRP still surpasses OE by +2.90\%. However, we want to clarify that comparing OE with RRP is not strictly fair. OOD regularization techniques like OE involve a distinct training regime, often needing an extra auxiliary OOD dataset. This is akin to unfairly providing one method with extra novel data while withholding such data from other methods. We present these empirical results as an additional point of reference for comparing RRP with existing OOD regularization techniques and not as an argument for the strict superiority of RRP.

\begin{table}[H]
\centering
\setlength\tabcolsep{2.5pt}
\renewcommand{\arraystretch}{1.25}
\scriptsize{
\begin{tabular}{c|c|cc|cc|cc|cc|cc}
\toprule
\multicolumn{1}{c|}{\multirow{2}{*}{\textbf{$\bm{D_{in}}$}}} & \multicolumn{1}{c|}{\multirow{2}{*}{\textbf{Method}}} & \multicolumn{2}{c|}{\textbf{iSUN}} & \multicolumn{2}{c|}{\textbf{LSUN Resize}} & \multicolumn{2}{c|}{\textbf{Places365}} & \multicolumn{2}{c|}{\textbf{Textures}} & \multicolumn{2}{c}{\textbf{Average}} \\ 
\cline{3-12} \multicolumn{1}{c|}{} & \multicolumn{1}{c|}{} &  FPR95 ($\downarrow$) & AUROC ($\uparrow$) & FPR95 ($\downarrow$) & AUROC ($\uparrow$) & FPR95 ($\downarrow$) & AUROC ($\uparrow$) & FPR95 ($\downarrow$) & AUROC ($\uparrow$) & FPR95 ($\downarrow$) & AUROC ($\uparrow$) \\
\midrule
\multirow{3}{*}{\textbf{CIFAR-10}}
& OE & \textbf{2.88} & \textbf{99.05} & \textbf{1.73} & \textbf{99.40} & \textbf{12.32} & \textbf{97.99} & \textbf{19.60} & \textbf{95.58} & \textbf{9.13} & \textbf{98.01} \\
[1ex] \cline{2-12} \rule{0pt}{3ex}
& RP  & 29.86$^{\pm 0.7}$ & 95.36$^{\pm 0.5}$ & 26.31$^{\pm 0.6}$ & 95.88$^{\pm 0.4}$ & 15.95$^{\pm 0.5}$ & 97.60$^{\pm 0.3}$ & 30.68$^{\pm 0.8}$ & 93.65$^{\pm 0.5}$ & 25.70$^{\pm 0.7}$ & 95.62$^{\pm 0.4}$\\
& RRP & 24.87$^{\pm 0.6}$ & 96.19$^{\pm 0.4}$ & 20.52$^{\pm 0.6}$ & 97.12$^{\pm 0.3}$ & 15.22$^{\pm 0.5}$ & 97.86$^{\pm 0.2}$ & 26.37$^{\pm 0.6}$ & 94.87$^{\pm 0.5}$ & 21.75$^{\pm 0.6}$ & 96.51$^{\pm 0.4}$\\
\bottomrule
\end{tabular}
}
\vspace{0.2cm}
\caption{\textbf{OOD Detection Results.} OOD detection performance comparison with outlier exposure (OE) and \texttt{CIFAR-10} as the in-distribution dataset. All values are percentages and \textbf{bold} values are the superior results.}
\label{tab:oe_ood_detection}
\end{table}

\section{Down-sampling Experiments}
\label{app:down_sampling}
In this appendix, we present our down-sampling experiments showcasing that our method isn't limited by the down-sampling step we implemented within our experiments. Specifically, in Table~\ref{tab:app_imagenet_generalization} we show the OOD generalization performance of \name{} and in Table~\ref{tab:app_imagenet_detection} we show the OOD detection performance of \name{} as we apply different degrees of down-sampling to training and testing datasets.

\begin{table}[H]
\vspace{0.1cm}
\centering
\vspace{0.2cm}
\setlength\tabcolsep{7pt} 
\renewcommand{\arraystretch}{1.1}
\scriptsize{
\begin{tabular}{c|c|c|c|c|c|c}
\toprule
\multirow{3}{*}{\textbf{\begin{tabular}[c]{@{}c@{}}Down-\\sampling\\Size\end{tabular}}} & \multirow{3}{*}{\textbf{\begin{tabular}[c]{@{}c@{}} Method \end{tabular}}} & \multicolumn{1}{c|}{\textbf{ImageNet-1k}} & \multicolumn{1}{c|}{\textbf{ImageNetV2}} & \multicolumn{1}{c|}{\textbf{ImageNet-A}} & \multicolumn{1}{c|}{\textbf{ImageNet-R}} & \multicolumn{1}{c}{\textbf{ImageNet-S}} \\ 
\cline{3-7} & \multicolumn{1}{c|}{} & Accuracy & Accuracy & Accuracy & Accuracy & Accuracy \\
& \multicolumn{1}{c|}{} & \multicolumn{1}{c|}{$\uparrow$} & \multicolumn{1}{c|}{$\uparrow$} & \multicolumn{1}{c|}{$\uparrow$} & \multicolumn{1}{c|}{$\uparrow$} & \multicolumn{1}{c}{$\uparrow$}\\ 
\midrule
\multirow{4}{*}{$\bm{64\times 64}$}
& Zero-shot & 50.632 & 44.22 & 6.12 & 34.547 & 28.556 \\
\cline{2-7} \rule{0pt}{3ex}
& Linear-probing & 65.322 & 53.61 & 5.627 & 36.19 & 29.354 \\
& Full Fine-tuning & \textbf{70.33} & \textbf{58.33} & 5.28 & 30.55 & 29.018 \\[1ex]
\cline{2-7} \rule{0pt}{3ex}
& Reprogrammer & 65.814 & 54.71 & \textbf{7.08} & \textbf{36.923} & \textbf{30.653} \\
\midrule
\multirow{4}{*}{$\bm{96\times 96}$}
& Zero-shot & 57.284 & 50.59 & 9.813 & 41.043 & 35.587 \\
\cline{2-7} \rule{0pt}{3ex}
& Linear-probing & 70.464 & 58.89 & 9.16 & 39.973 & 35.517 \\
& Full Fine-tuning & \textbf{72.228} & \textbf{60.68} & 5.60 & 31.9 & 32.249 \\[1ex]
\cline{2-7} \rule{0pt}{3ex}
& Reprogrammer & 70.244 & 59.27 & \textbf{11.00} & \textbf{41.777} & \textbf{36.906} \\
\midrule
\multirow{4}{*}{$\bm{128\times 128}$}
& Zero-shot & 59.44 & 52.79 & 11.82 & 43.48 & 38.61 \\
\cline{2-7} \rule{0pt}{3ex}
& Linear-probing & 72.43 & 61.35 & 10.71 & 41.58 & 38.19 \\
& Full Fine-tuning & \textbf{73.14} & 60.98 & 6.41 & 32.71 & 32.83 \\[1ex]
\cline{2-7} \rule{0pt}{3ex}
& Reprogrammer & 72.10 & \textbf{61.28} & \textbf{12.58} & \textbf{44.30} & \textbf{39.40} \\

\midrule
\multirow{4}{*}{$\bm{160\times 160}$}
& Zero-shot & 60.14 & 53.46 & 12.893 & 44.36 & 39.932 \\
\cline{2-7} \rule{0pt}{3ex}
& Linear-probing & 73.142 & 61.86 & 11.787 & 42.167 & 39.27 \\
& Full Fine-tuning & \textbf{73.39} & 61.59 & 6.227 & 32.877 & 33.331 \\[1ex]
\cline{2-7} \rule{0pt}{3ex}
& Reprogrammer & 72.934 & \textbf{61.91} & \textbf{14.07} & \textbf{44.43} & \textbf{40.41} \\

\midrule
\multirow{4}{*}{$\bm{192\times 192}$}
& Zero-shot & 60.796 & 53.84 & 13.8 & 44.86 & 40.294 \\
\cline{2-7} \rule{0pt}{3ex}
& Linear-probing & 73.486 & 62.03 & 11.907 & 42.093 & 39.425 \\
& Full Fine-tuning & \textbf{73.764} & 61.86 & 6.427 & 32.137 & 32.135 \\[1ex]
\cline{2-7} \rule{0pt}{3ex}
& Reprogrammer & 73.08 & \textbf{62.36} & \textbf{14.73} & \textbf{44.57} & \textbf{40.73} \\
\midrule
\multirow{4}{*}{$\bm{224\times 224}$}
& Zero-shot & 61.896 & 54.71 & 15.267 & 46.713 & 40.83 \\
\cline{2-7} \rule{0pt}{3ex}
& Linear-probing & 74.882 & 62.45 & 12.6 & 42.217 & 39.944 \\
& Full Fine-tuning & \textbf{75.071} & 62.03 & 6.387 & 32.48 & 33.469 \\[1ex]
\cline{2-7} \rule{0pt}{3ex}
& Reprogrammer & 74.118 & \textbf{62.65} & \textbf{15.32} & \textbf{45.09} & \textbf{40.86} \\
\bottomrule
\end{tabular}
}
\caption{\textbf{Down-sampling OOD Generalization Results} OOD generalization performance comparison between differing down-sampling severity. All methods utilize the CLIP B/32 architecture and were fine-tuned on the \texttt{ImageNet-1k} dataset down-sampled to the specified resolution. Similarly, the evaluation was completed using, if available, the validation dataset down-sampled to the specified resolution. $\uparrow$ indicates larger values are better, while $\downarrow$ indicates smaller values are better. All values are percentages and \textbf{bold} numbers are superior \textbf{fine-tuning} results.}
\label{tab:app_imagenet_generalization}
\end{table}

\newpage

\begin{table*}[t]
\centering
\setlength\tabcolsep{5.15pt}
\renewcommand{\arraystretch}{1.25}
\scriptsize{
\begin{tabular}{c|c|cc|cc|cc|cc|cc}
\toprule
\multirow{3}{*}{\textbf{\begin{tabular}[c]{@{}c@{}}Down-\\sampling\\Size\end{tabular}}} & \multicolumn{1}{c|}{\multirow{3}{*}{\textbf{Method}}} & \multicolumn{2}{c|}{\textbf{iNaturalist}} & \multicolumn{2}{c|}{\textbf{SUN}} & \multicolumn{2}{c|}{\textbf{Places}} & \multicolumn{2}{c|}{\textbf{Textures}} & \multicolumn{2}{c}{\textbf{Average}} \\ 
\cline{3-12} \multicolumn{1}{c|}{} & \multicolumn{1}{c|}{} &  FPR95 & AUROC & FPR95 & AUROC & FPR95 & AUROC & FPR95 & AUROC & FPR95 & AUROC \\
\multicolumn{1}{c|}{} & \multicolumn{1}{c|}{} & \multicolumn{1}{c}{$\downarrow$} & \multicolumn{1}{c|}{$\uparrow$} & \multicolumn{1}{c}{$\downarrow$} & \multicolumn{1}{c|}{$\uparrow$} & \multicolumn{1}{c}{$\downarrow$} & \multicolumn{1}{c|}{$\uparrow$} & \multicolumn{1}{c}{$\downarrow$} & \multicolumn{1}{c|}{$\uparrow$} & \multicolumn{1}{c}{$\downarrow$} & \multicolumn{1}{c}{$\uparrow$} \\
\midrule
\multirow{4}{*}{$\bm{64\times 64}$}
& ZS & 63.57 & 81.3 & 75.1 & 77.8 & 74.34 & 76.14 & 73.67 & 75.82 & 71.67 & 77.77 \\
[1ex] \cline{2-12} \rule{0pt}{3ex}
& LP & 65.44 & \textbf{84.62} & 84.61 & 70.23 & 83.48 & 70.79 & \textbf{76.35} & \textbf{76.51} & 77.47 & 75.54 \\
& FFT & 74.82 & 79.44 & 81.29 & 73 & 80.55 & \textbf{73.45} & 80.96 & 72.95 & 79.4 & 74.71 \\
[1ex] \cline{2-12} \rule{0pt}{3ex}
& RP  & \textbf{65.42} & 83.68 & \textbf{79.96} & \textbf{72.05} & \textbf{80.13} & 71.86 & 77.78 & 74.72 & \textbf{75.82} & \textbf{75.58}\\
\midrule
\multirow{4}{*}{$\bm{96\times 96}$}
& ZS & 57.5 & 83.7 & 67.13 & 80.53 & 68 & 78.77 & 71.13 & 76.64 & 65.94 & 79.91 \\
[1ex] \cline{2-12} \rule{0pt}{3ex}
& LP & \textbf{54.21} & \textbf{87.46} & 80.32 & 73.77 & 77.94 & \textbf{74.18} & \textbf{72.36} & \textbf{78.28} & 71.21 & \textbf{78.42} \\
& FFT & 73.33 & 80.55 & 80.89 & 73.43 & 80.03 & 74 & 80.09 & 73.75 & 78.58 & 75.43 \\
[1ex] \cline{2-12} \rule{0pt}{3ex}
& RP  & 56.76 & 85.62 & \textbf{77.19} & \textbf{73.86} & \textbf{76.46} & 73.91 & 73.07 & 76.78 & \textbf{70.87} & 77.54\\
\midrule
\multirow{4}{*}{$\bm{128\times 128}$}
& ZS & 53.96 & 85.15 & 64.89 & 81.26 & 65.76 & 79.30 & 70.05 & 77.03 & 63.67 & 80.69 \\
[1ex] \cline{2-12} \rule{0pt}{3ex}
& LP & \textbf{51.15} & \textbf{88.25} & 78.68 & 74.58 & 76.42 & \textbf{75.15} & \textbf{70.25} & \textbf{78.71} & \textbf{69.12} & \textbf{79.17} \\
& FFT & 71.94 & 81.37 & 80.29 & 74.01 & 79.97 & 74.54 & 78.28 & 74.80 & 77.62 & 76.18 \\
[1ex] \cline{2-12} \rule{0pt}{3ex}
& RP  & 56.85 & 85.97 & \textbf{75.68} & \textbf{74.99} & \textbf{74.80} & 74.84 & 70.51 & 77.43 & 69.46 & 78.31 \\
\midrule
\multirow{4}{*}{$\bm{160\times 160}$}
& ZS & 52.98 & 85.38 & 63.57 & 81.5 & 64.36 & 79.63 & 69.34 & 77.23 & 62.56 & 80.93 \\
[1ex] \cline{2-12} \rule{0pt}{3ex}
& LP & \textbf{50.42} & \textbf{88.37} & 77.53 & 74.94 & 75.17 & 75.65 & 68.55 & \textbf{79.01} & 67.92 & \textbf{79.49} \\
& FFT & 71.83 & 81.15 & 81.55 & 73.26 & 80.35 & 74.2 & 78.79 & 74.43 & 78.13 & 75.76 \\
[1ex] \cline{2-12} \rule{0pt}{3ex}
& RP  & 56.25 & 85.92 & \textbf{74.55} & \textbf{75.69} & \textbf{72.37} & \textbf{76.27} & \textbf{67.98} & 77.93 & \textbf{67.79} & 78.95 \\
\midrule
\multirow{4}{*}{$\bm{192\times 192}$}
& ZS & 53.57 & 85.22 & 63.75 & 81.37 & 63.04 & 80.16 & 68.99 & 77.44 & 62.34 & 81.05 \\
[1ex] \cline{2-12} \rule{0pt}{3ex}
& LP & \textbf{50.28} & \textbf{88.36} & 78.02 & 74.89 & 74.62 & 76.25 & \textbf{69.04} & \textbf{78.96} & 67.99 & \textbf{79.62} \\
& FFT & 71.95 & 81.09 & 81.43 & 73.56 & 81.11 & 74.12 & 79.2 & 74.6 & 78.42 & 75.84 \\
[1ex] \cline{2-12} \rule{0pt}{3ex}
& RP  & 55.77 & 85.99 & \textbf{74.48} & \textbf{75.31} & \textbf{71.30} & \textbf{76.79} & 69.24 & 77.8 & \textbf{67.70} & 78.97 \\
\midrule
\multirow{4}{*}{$\bm{224\times 224}$}
& ZS & 53.75 & 85.58 & 62.89 & 81.65 & 63.82 & 80.13 & 68.19 & 77.67 & 62.16 & 81.26 \\
[1ex] \cline{2-12} \rule{0pt}{3ex}
& LP & \textbf{50.88} & \textbf{88.18} & 79.14 & 74.92 & 75.9 & 76.02 & \textbf{67.73} & \textbf{79.35} & 68.41 & \textbf{79.62} \\
& FFT & 72.14 & 80.89 & 80.98 & 74.06 & 80.69 & 74.58 & 79.04 & 74.98 & 78.21 & 76.13 \\
[1ex] \cline{2-12} \rule{0pt}{3ex}
& RP  & 55.56 & 85.82 & \textbf{73.89} & \textbf{76.25} & \textbf{70.32} & \textbf{77.63} & 68.05 & 78.28 & \textbf{66.95} & 79.5 \\
\bottomrule
\end{tabular}
}
\caption{\textbf{Down-sampling OOD Detection Results} OOD detection performance comparison between differing down-sampling severity. All methods utilize the CLIP B/32 architecture fine-tuned on the \textbf{Image-1k} dataset down-sampled to the specified resolution. Similarly, all semantically shifted OOD datasets were also down-sampled to the specified resolution. $\uparrow$ indicates larger values are better, while $\downarrow$ indicates smaller values are better. All values are percentages and \textbf{bold} numbers are the superior \textbf{fine-tuning} results.}
\vspace{-0.1cm}
\label{tab:app_imagenet_detection}
\end{table*}

\end{document}